\documentclass[10pt,twocolumn,letterpaper]{article}
\pdfoutput=1

\usepackage{cvpr}              
\makeatletter
\@namedef{ver@everyshi.sty}{}
\makeatother

\usepackage[accsupp]{axessibility}

\newcommand{\tightpara}[1]{\vspace*{-3mm}\paragraph{#1}}

\newcommand{\final}{0}  
\usepackage{geometrycollective}

\def\PaperTitle{Understanding and Improving Features Learned in Deep Functional Maps}

\theoremstyle{definition}
\newtheorem{definition}{Definition}[section]

\newtheorem{theorem}{Theorem}[section]

\newtheorem{innercustomthm}{Theorem}
\newenvironment{customthm}[1]
  {\renewcommand\theinnercustomthm{#1}\innercustomthm}
  {\endinnercustomthm}

\newcommand{\C}{\mathbf{C}}
\newcommand{\A}{\mathbf{A}}

\newcommand{\opt}{\rm{opt}}

\def\cvprPaperID{1770} 
\def\confName{CVPR}
\def\confYear{2023}

\begin{document}

\title{\PaperTitle}

\author{Souhaib Attaiki\\
LIX, École Polytechnique, IP Paris\\
{\tt\small attaiki@lix.polytechnique.fr}
\and
Maks Ovsjanikov\\
LIX, École Polytechnique, IP Paris\\
{\tt\small maks@lix.polytechnique.fr}
}
\maketitle


\begin{abstract}
Deep functional maps have recently emerged as a successful paradigm for non-rigid 3D shape correspondence tasks. An essential step in this pipeline consists in learning feature functions that are used as constraints to solve for a functional map inside the network. However, the precise nature of the information learned and stored in these functions is not yet well understood. Specifically, a major question is whether these features can be used for any other objective, apart from their purely algebraic role in solving for functional map matrices. In this paper, we show that under some mild conditions, the features learned within deep functional map approaches can be used as point-wise descriptors and thus are directly comparable across different shapes, even without the necessity of solving for a functional map at test time. Furthermore, informed by our analysis, we propose effective modifications to the standard deep functional map pipeline, which promote structural properties of learned features, significantly improving the matching results. Finally, we demonstrate that previously unsuccessful attempts at using extrinsic architectures for deep functional map feature extraction can be remedied via simple architectural changes, which encourage the theoretical properties suggested by our analysis. We thus bridge the gap between intrinsic and extrinsic surface-based learning, suggesting the necessary and sufficient conditions for successful shape matching. Our code is available at  \url{https://github.com/pvnieo/clover}.
\end{abstract}

\section{Introduction}
\label{sec:intro}

Computing dense correspondences between 3D shapes is a classical problem in Geometry Processing, Computer Vision, and related fields, and remains at the core of many tasks including statistical shape analysis \cite{Bogo2014,Pishchulin2017}, registration \cite{Zhou2016}, deformation  \cite{Baran2009} or texture transfer \cite{Dinh2005} among others.

Since its introduction, the functional map (fmap) pipeline \cite{ovsjanikov2012functional} has become a de facto tool for addressing this problem. This framework relies on representing correspondences as linear transformations across functional spaces, by encoding them as small matrices using the Laplace-Beltrami basis. Methods based on this approach have been successfully applied with hand-crafted features \cite{Salti2014,sun2009concise,aubry2011wave} to many scenarios including near-isometric, \cite{Ren2019, huang2014functional,eynard2016coupled,Shoham2019,Melzi_2019}, non-isometric \cite{kovnatsky2013coupled,Eisenberger2020SmoothSM} and partial \cite{cosmo2016shrec,Litany2017,Litany2016,Xiang_2021_CVPR,Wu2020,Rodol2016} shape matching. In recent years, a growing body of literature has  advocated improving the functional map pipeline by using \textit{deeply} learned features, pioneered by \cite{litany2017deep}, and built upon by many follow-up works \cite{donati2020deep,sharma2020weakly,attaiki2021dpfm,Marin2020CorrespondenceLV,eisenberger2020deep,marin22_why,halimi2019unsupervised,sharp2021diffusion}. In all of these methods, the learned features are only used to constrain the linear system when estimating the functional maps inside the network. Thus, no attention is paid to their geometric nature, or potential utility beyond this purely algebraic role.

\begin{figure}
    \centering
    \includegraphics[width=\columnwidth]{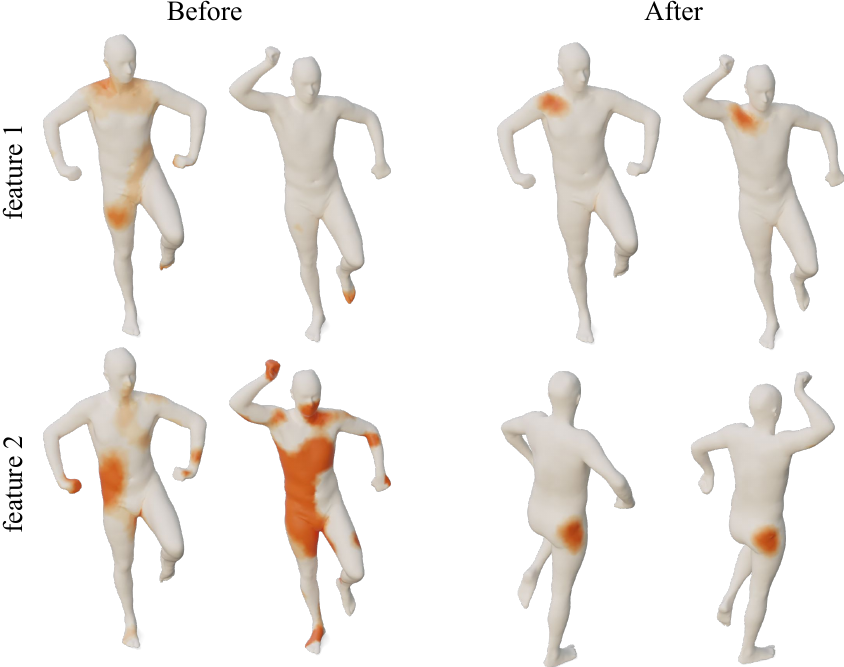}
    \caption{\textit{Left:} Features learned in existing deep functional map pipelines are used in a purely \textit{algebraic manner} as constraints for a linear system, thus lacking interpretability and clear geometric content. \textit{Right:} We propose a theoretically-justified modification to this pipeline, which leads to learning \textit{robust} and \textit{repeatable} features that enable matching via nearest neighbor search.}
    \label{fig:teaser}
    \vspace{\figmargin}
\end{figure}

On the other hand, features learned in other deep matching paradigms are the main focus of optimization, and they represent either robust descriptive geometric features that are used directly for matching using nearest neighbor search \cite{litman2013learning,Zeng_2017_CVPR,Deng_2018_CVPR,Gojcic_2019_CVPR,Yew_2020_CVPR,bai2021pointdsc,li2022srfeat}, or as distributions that, at every point, are interpreted as vertex indices on some template shape \cite{masci2015geodesic,monti2017geometric,poulenard2018multi}, or a deformation field that is used to deform the input shape to match a template \cite{groueix20183d}. 

In contrast, feature (also known as ``probe'' \cite{Ovsjanikov2017}) functions, within deep functional maps are used purely as an optimisation tool and thus the information learned and stored in these functions is not yet well-understood. In this work, we aim to show that features in deep functional maps networks can, indeed, have geometric significance and that, under certain conditions, they are directly comparable across shapes, and can be used for matching simply via nearest neighbor search, see \cref{fig:teaser}.

Specifically, we introduce the notion of feature completeness and show that under certain mild conditions, extracting a pointwise map from a functional map or via the nearest neighbor search between learned features leads to the same result. Secondly, we propose a modification of the deep functional map pipeline, by imposing the learned functional maps to satisfy the conditions imposed by our analysis. We show that this leads to a significant improvement in accuracy, allowing state-of-the-art results by simply performing a nearest-neighbor search between features at test time. Finally, based on our theoretical results, we also propose a modification of some extrinsic feature extractors \cite{dgcnn,Wiersma2022DeltaConv}, which previously failed in the context of deep functional maps, which improve their overall performance by a significant margin. Since our theoretical results hold for the functional map paradigm in general, they can be incorporated in \emph{any} deep fmap method, hence improving previous methods, and making future methods more robust.

Overall, our contributions can be summarized as follows:

\begin{itemize}[topsep=2pt,partopsep=2pt,itemsep=1pt,parsep=3pt]
    \item We introduce the notions of feature completeness and basis aligning functional maps and use them to establish a theoretical result about the nature of features learned in the deep functional map framework.
    \item Informed by our analysis, we propose simple modifications to this framework, which lead to state-of-the-art results in challenging scenarios.
    \item Based on our theoretical results, we propose a simple modification to some extrinsic feature extractors, that were previously unsuccessful for deep functional maps, improving their overall accuracy, and bridging the gap between intrinsic and extrinsic surface-based learning.
\end{itemize}

\section{Related Work}
\label{sec:related}

Non-rigid shape matching is a very rich and well-established research area, and a full overview is beyond
the scope of this paper.  Below, we review works that are close to our method and  refer the interested reader to recent surveys \cite{cao2020comprehensive,bronstein2017geometric,guo2016comprehensive,guo2020deep} for a more in-depth treatment.

\tightpara{Functional Maps}
Since its introduction \cite{Ovsjanikov2012}, the functional map pipeline has become a widely-used tool for non-rigid shape matching, being adopted and extended in many follow-up works \cite{ginzburg2019cyclic,Ren2019,eynard2016coupled,Melzi_2019,eynard2016coupled,Nogneng2017,rodola2017partial,poulenard_persistence,sharma2020weakly}. The advantage of this approach is that it reduces the optimization of pointwise maps, which are quadratic in the number of vertices, to optimizing small matrices, thus greatly simplifying computational problems. We refer to \cite{Ovsjanikov2017} for an overview. 

The original functional map pipeline relied on input feature (probe) functions, which were given a priori \cite{Salti2014,sun2009concise,aubry2011wave}.
Subsequent research has improved the method by adapting it to partial shapes \cite{cosmo2016shrec,rodola2017partial,Litany2017} or using robust regularizers  
\cite{Ren2019,Nogneng2017,kovnatsky2013coupled,burghard2017embedding} and proposed efficient refinement techniques \cite{Melzi_2019,Pai_2021_CVPR,jing_maptree}. 
In all of these works, fmaps were computed using hand-crafted probe functions, and any information loss in these descriptors hinders downstream optimization.

More recent works have proposed to solve this problem by learning descriptors (probe functions) directly from the data, using deep neural networks. This line of research was initiated by FMNet \cite{litany2017deep}, and extended in many subsequent works \cite{halimi2019unsupervised,attaiki2023vader,roufosse2019unsupervised,sharma2020weakly,sharp2021diffusion,attaiki2021dpfm,donati2020deep,Eisenberger2021NeuroMorphUS,attaiki2022ncp,eisenberger2020deep,li2022srfeat}. These methods learn the probe features directly from the raw geometry, using a neural network, and supervise the learning with a loss on the functional map in the reduced basis.


A parallel line of work has focused on making the learning \textit{unsupervised}, which can be convenient in the absence of ground truth correspondences. Approaches in \cite{halimi2019unsupervised} and \cite{Ginzburg2020} have proposed penalizing either the geodesic distortion of the pointwise map predicted by the network or using the cycle consistency loss.  
Another line of work \cite{roufosse2019unsupervised,sharma2020weakly,sharp2021diffusion,donati-duo} considered unsupervised training by imposing structural properties on the functional maps in the reduced basis, such as bijectivity, orthonormality, and commutativity of the Laplacian. 
The authors of \cite{sharma2020weakly} have shown that feature learning can be done starting from raw 3D geometry in the weakly supervised setting, where shapes are only approximately rigidly pre-aligned. 

In all of these works, features extracted by neural networks have been used to formulate the optimization problem, from which a functional map is computed. Thus, 
so far no attention has been paid to the geometric nature or other utility of learned probe functions. In contrast, we aim to analyze the conditions under which probe functions can be used for direct pointwise map computation and use this analysis to design improved map estimation pipelines. 

We also note briefly a very recent work  \cite{li2022srfeat} which has advocated for imposing feature smoothness  when learning for non-rigid shape correspondence. However, that work does not use the in-network functional map estimation and moreover lacks any theoretical analysis or justification for its design choices. 
%

\tightpara{Recent Advances in Axiomatic Functional Maps}
Our results are also related to recent axiomatic methods for functional map-based methods, which  \textit{couple} optimization for the functional map (fmap) with the associated point map (p2p map) \cite{ren2018continuous,Melzi_2019,Pai_2021_CVPR,Xiang_2021_CVPR}. These techniques typically propose to refine functional maps by \textit{iterating} the conversion from functional to pointwise maps and vice versa. This approach was recently summarized in \cite{discrete_Ren2021}, where the authors introduce the notion of functional map ``properness'' and describe a range of energies that can be optimized via this iterative conversion scheme. 

The common denominator between all these methods is the inclusion of pointwise maps in the process of functional map optimization. In this work, we propose a method and a loss that similarly incorporate pointwise map computation. However, we do so in a learning context and show that this leads to significant improvements in the overall accuracy of the deep functional map pipeline.

\tightpara{Learning on Surfaces}
Multiple methods for deep surface learning have been proposed to address the limitations of handcrafted features in downstream tasks. One type of method is point-based (extrinsic) methods, such as PointNet \cite{qi2017pointnet} and follow-up works \cite{qi2017pointnet++,rethink_ma_22,thomas2019KPConv,dgcnn,pcnn_2018,Wiersma2022DeltaConv}. These methods are simple, effective, and widely applicable. However, they often fail to generalize to new datasets or significant pose changes in deep shape matching. 

Another line of research \cite{poulenard2018multi,sharp2021diffusion,wiersma2020cnns,gong2019spiralnet++,masci2015geodesic,maron2017convolutional} (intrinsic methods) has focused on defining the convolution operator directly on the surface of the input shape. These methods are more suitable for deformable shapes and can leverage the structure of the surface encoded by the mesh, which is ignored by the extrinsic methods.



In previous works, intrinsic methods tend to perform better for non-rigid shape matching than extrinsic methods, with DiffusionNet \cite{sharp2021diffusion} being considered the state-of-the-art feature extractor for shape matching. In this work, we will show that a simple modification to extrinsic feature extractors improves their overall performance for shape matching, making them comparable to or better than DiffusionNet. 

\paragraph{Paper Organization}
The rest of the paper is structured as follows. In \cref{sec:background}, we first introduce the deep functional map pipeline and our key definitions. We then state a theorem that shows that features learned in the deep functional map pipeline have a geometric interpretation under certain conditions. In light of this discovery, we introduce in \cref{sec:method} two modifications to the functional map pipeline, by promoting structural properties suggested by our analysis. Finally, in \cref{sec:application}, we will show how these modifications improve overall performance in different shape matching scenarios.
\section{Notation, Background \& Motivation}
\label{sec:background}

\subsection{Notation}
\label{sec:notation}

Suppose we are given a pair of shapes $S_1, S_2$ represented as triangle meshes with respectively $n_1$ and $n_2$ vertices. We compute the cotangent Laplace-Beltrami operator \cite{Meyer2003} of each shape $S_i$ and collect the first k eigenfunctions as columns in a matrix denoted by $\Phi_i$, and the corresponding eigenvalues in a diagonal matrix denoted as $\Delta_i$. $\Phi_i$ is orthonormal with respect to the area (mass) matrix $\Phi_i^{T} A_i \Phi_i = \mathbb{I}$. We denote $\Phi_i^{\dagger} = \Phi_i^{T} A_i$, where $\cdot^{\dagger}$ is the (left) Moore–Penrose pseudo-inverse.

A pointwise map $T_{12}: S_1 \rightarrow S_2$ can be encoded in a matrix $\Pi_{12} \in \mathbb{R}^{n_1 \times n_2},$ where $\Pi_{12}(i,j) = 1$ if $T_{12}(i) = j$ and 0 otherwise. We will use the letter $\Pi$ to denote pointwise maps in general. To every pointwise map, one can associate a functional map, simply via projection: $\C_{21} = \Phi^{\dagger}_1 \Pi_{12} \Phi_{2}$.

Let $\mathcal{F}_{\Theta}$ be some feature extractor, which takes as input a shape and produces a set of feature functions (where $\Theta$ are some trainable parameters). We then have $F_1 = \mathcal{F}_{\Theta}(S_1)$. For simplicity we will omit explicitly stating $\mathcal{F}_{\Theta}$ and just denote the features associated with a shape $S_i$ by $F_i$, where the presence of some feature extractor is assumed implicitly. Finally, $\A_i = \Phi_i^{\dagger} F_i$ denotes the matrix of coefficients of the feature functions in the corresponding reduced basis.

\subsection{Deep Functional Map Pipeline}

The standard Deep Functional Map pipeline can be described as follows.
     \tightpara{Training:} Pick a pair of shapes $S_1, S_2$ from some training set $\{S_i\}$, and compute their feature functions $F_1 = \mathcal{F}_{\Theta}(S_1), F_2 = \mathcal{F}_{\Theta}(S_2)$. Let $\A_{1} = \Phi_1^{\dagger} F_1$, and $\A_{2} = \Phi_2^{\dagger} F_2$ denote the coefficients of the feature functions in the reduced basis.
       Compute the functional map $\C_{12}$ by solving a least squares system inside the network:
        \begin{align}
            \argmin_{\C} \| \C \A_1 - \A_2 \|_F^2.
            \label{eq:fmap_basic}
        \end{align}
    This system can be further regularized by incorporating commutativity with the Laplacian \cite{donati2020deep}. A \emph{training loss} is then imposed on this computed functional map (in the supervised setting) by comparing the computed functional map $\C_{12}$ with some ground truth:
        \begin{align}
            \mathcal{L}_{sup}(\C_{12}) = \| \C_{12} - \C_{\rm{gt}} \|_F^2. \label{eq:sup_loss}
        \end{align}
        Alternatively, (in the unsupervised setting) we can impose a loss on $\C_{12}$ by penalizing its deviation from some desirable properties, such as being an orthonormal matrix \cite{roufosse2019unsupervised}. Finally, the loss above is used to back-propagate through the feature extractor network $\mathcal{F}_{\Theta}$ and optimize its parameters $\Theta$.

   \tightpara{Test time:} Once the parameters of the feature extractor network are trained, we follow a similar pipeline at test time. Given a pair of \emph{test shapes} $S_1, S_2$, we compute their feature functions $F_1 = \mathcal{F}_{\Theta}(S_1), F_2 = \mathcal{F}_{\Theta}(S_2)$. Let $\A_{1} = \Phi_1^{\dagger} F_1$, and $\A_{2} = \Phi_2^{\dagger} F_2$ denote their coefficients in the reduced basis. We then compute the functional map $\C_{12}$ by solving the least squares system in \cref{eq:fmap_basic}. Given this functional map, a point-to-point (p2p) correspondence $\Pi_{21}$ can be obtained by solving the following problem:
        \begin{align}
            \Pi_{21} = \argmin_{\Pi} \| \Pi \Phi_1 - \Phi_2 \C_{12}\|.
            \label{eq:fmap_conversion}
        \end{align}
        This problem, which was proposed and justified in \cite{ezuz2017deblurring,Pai_2021_CVPR} is row-separable and reduces to the nearest neighbor search between the rows of $\Phi_1$ and the rows of $\Phi_2 \C_{12}$. This method of converting the fmap into a p2p map is referred to hereafter as the \textit{adjoint method} following \cite{Pai_2021_CVPR}.
  
\vspace{-0.7em}
\paragraph{Advantages}
This pipeline has several advantages: first, the training loss is imposed on the entire map rather than individual point correspondences, which has been shown to improve accuracy \cite{litany2017deep,donati2020deep}. Second, by using a reduced spectral basis, this approach strongly regularizes the learning problem and thus can be used even in the presence of limited training data \cite{donati2020deep,sharp2021diffusion}. Finally, by accommodating both supervised and unsupervised losses, this pipeline allows to promote structural map properties without manipulating large and dense, e.g., geodesic distance, matrices.


\vspace{-0.7em}
\paragraph{Drawbacks \& Motivation}
At the same time, conceptually, the relation between the learned feature functions and the computed correspondences is still unclear. Indeed, although the feature (probe) functions are learned, they are used solely to formulate the optimization problem in \cref{eq:fmap_basic}. Moreover, the final pointwise  correspondence in \cref{eq:fmap_conversion} is still computed from the Laplacian basis. Thus, it is not entirely clear what exactly is learned by the network $\mathcal{F}_\Theta$, and whether the learned feature functions can be used for any other task. Indeed, in practice, learned features can fail to highlight some specific repeatable regions on shapes, even if they tend to produce high-quality functional map matrices (\cref{fig:teaser}).

\subsection{Theoretical Analysis}
\label{sec:motivation}
In this subsection, we introduce some notions and state a theorem that will be helpful in our results below. We will use the same notation as in \cref{sec:notation}.

First, note that both during training (in \cref{eq:fmap_basic}) and at test time (in  \cref{eq:fmap_conversion}) the learned feature functions $F_i$, are used by projecting them onto the Laplacian basis. Thus, both from the perspective of the loss and also when computing the pointwise map, the deep functional map pipeline only uses the part of the functions that lies in the span of the Laplacian basis. Put differently, if we \textit{modified} feature functions by setting $\tilde{F}_i  = \Phi_i \Phi_i^{\dagger} F_i$ then both during training and at test time the behavior of the deep functional map pipeline will be identical when using either $F_i$ or $\tilde{F}_i$.

\begin{definition}[Complete feature functions]
Motivated by this observation, we call the feature extractor $F_{\Theta}$ \textbf{complete} if $F_{\Theta}$ produces feature (probe) functions that are contained in the corresponding LB eigenbasis. I.e.,
\begin{align}
	&\mathcal{F}_\Theta (S_i) = \Phi_i \Phi_i^{\dagger} \mathcal{F}_\Theta (S_i) , \text{ or equivalently} \\
	&(Id - \Phi_i \Phi_i^{\dagger}) \mathcal{F}_\Theta (S_i)  = 0.
\end{align}
\end{definition}
%
%



\begin{definition}[Basis aligning functional maps] We also introduce a notion that relates to the properties of functional maps. Namely, given a pair of shapes, $S_1, S_2$ and a functional map $\C_{12}$, we will call it \textit{basis-aligning} if the bases, when transformed by this functional map, align exactly. This can be summarized as follows: $\Phi_{2} \C_{12} = \Pi_{21} \Phi_{1}$ for some point-to-point map $\Pi_{21}$.
\end{definition}

For simplicity, for our result below we will also assume that all optimization problems have unique global minima. Thus, for the problem $\argmin_{\C} \|\C \A_{1} - \A_{2} \|$, this means that $\A_1$ must be full rank, whereas for the problem of type $\argmin_{\Pi} \| \Pi F_1 - F_2 \|$, this means that the rows of $F_1$ must be distinct (i.e., no two rows are identical, as vectors).




\begin{theorem}
\label{thm:equivalence}
    Suppose the feature extractor $F_{\Theta}$ is \emph{complete}. Let $\A_1 = \Phi_1^{\dagger} F_1$ and $\A_2 = \Phi_2^{\dagger} F_2$. Then, denoting $\C_{\rm{opt}} = \argmin_{\C} \|\C \A_{1} - \A_{2} \|$, we have the following results hold:
    
(1) If $\Pi F_1 = F_2$ for some point-to-point map $\Pi$ then $\C_{12} = \Phi_2^{\dagger} \Pi \Phi_1$ is basis-aligning. Moreover, $\C_{12} = \C_{\opt}$, and extracting the pointwise map from $\C_{\opt}$ via the adjoint method (see \cref{eq:fmap_conversion}) or via nearest neighbor search in the feature space $\min_{\Pi} \| \Pi F_1 - F_2 \|$ will give the same result.

(2) Conversely, suppose that $F_1, F_2$ are complete and $\C_{\opt}$ is basis aligning, then $\argmin_{\Pi} \| \Pi F_1 - F_2 \| = \argmin_{\Pi} \| \Pi \Phi_1 - \Phi_2 C_{\opt}\|$.
\end{theorem}

\begin{proof}
    See the supplementary materials.
\end{proof}

\tightpara{Discussion:} 
Note that in the theorem above, we used the notion of basis-aligning functional maps. Two questions arise: what are the necessary and sufficient conditions for a functional map to be basis-aligning? 

A necessary condition is that the functional map must be \textit{proper} as defined in \cite{ren2021discrete}. Namely, a functional map $\C_{12}$ is \textit{\textbf{proper}} if there exists a pointwise map $\Pi_{21}$ such that  $\C_{12} = \Phi_{2}^{\dagger} \Pi_{21} \Phi_{1}$.  Note that for sufficiently high dimensionality $k$, \textit{any} proper functional map must be basis aligning.  This is because, since $\Phi_{2}$ is guaranteed to be invertible in the full basis, there is a unique matrix $\C_{12}$ that will satisfy $\Pi_{21} \Phi_{1} = \Phi_{2} \C_{12}$ exactly.

Conversely, a \textit{sufficient} condition for a functional map to be basis aligning is that the underlying point-to-point map is an isometry. Indeed, it is well known that isometries must be represented as block-diagonal functional maps, and moreover, isometries preserve eigenfunctions (see Theorem 2.4.1 in \cite{Ovsjanikov2017} and Theorem 1 in \cite{rustamov2013map}). Thus, assuming that the functional map is of size $k$ and that the $k^{\text{th}}$ and $(k+1)^{\text{st}}$ eigenfunctions are distinct, we must have that the functional map must be basis-aligning.

Finally, note that a functional map is basis-aligning if it is \textit{proper} and if the image (by pull-back) of the $k$ first eigenfunctions of the source lies in the range of the $k$ first eigenfunctions of the target. This can therefore be considered as a measure of the \textit{smoothness} of the map.

%
%
%

\section{Proposed Modification}
\label{sec:method}

In the previous section, we provided a theoretical analysis of the conditions under which computed ``probe'' functions within the deep functional map pipeline can be used as pointwise descriptors directly, and lead to the same point-to-point maps as computed by the functional maps. In \cref{sec:application}, we provide an extensive evaluation of using learned feature functions for pointwise map computation and thus affirm the validity of \cref{thm:equivalence} in practice.

Our main observation is that the two approaches for point-to-point map computation are indeed often equivalent in practice, especially in ``easy'' cases, where existing state-of-the-art approaches lead to highly accurate maps. In contrast, we found that in more challenging cases, where existing methods fail, the two approaches are not equivalent and can lead to significantly different results. 


Motivated by our analysis, we propose to use the structural properties suggested in \cref{thm:equivalence} as a way to bridge this gap and improve the overall accuracy. Our proposed modifications are relatively simple, but they significantly improve the quality of computed correspondences, especially in ``difficult'' matching scenarios, as we demonstrate in \cref{sec:application}.


The two key assumptions in \cref{thm:equivalence} are \textit{basis-aligning} functional maps and \textit{complete} feature extractors. We propose modifying the functional map pipeline to satisfy the conditions of the theorem. Since the basis-aligning property is closely related to \textit{properness}, we propose to impose that the predicted functional map to be proper, \ie arises from some pointwise correspondence. For feature completeness, we suggest modifying the feature extractor to produce \textit{smooth features}. We use the same notation as in \cref{sec:notation}.

\subsection{Enforcing Properness}
\label{sec:proper_fmap}

In this section, we propose two ways to enforce functional map properness and associated losses  for both supervised and unsupervised training.

\paragraph{The adjoint method}
Given feature functions $F_1, F_2$, produced by a feature extractor, we compute the functional map $\C_{12-pred}$ as explained in \cref{sec:background}. To compute a proper functional from it, we first convert $\C_{12-pred}$ into a p2p map $\Pi_{21-pred}$ in a differentiable way and then compute the ``differentiable'' proper functional map $\C_{21-proper} = \Phi_{2}^{\dagger} \Pi_{21-pred} \Phi_{1}$. 

To compute $\Pi_{21-pred}$, denoting $G_1 = \Phi_{1}$ and $G_2 = \Phi_{2}\C_{21-pred}$, we use:
\begin{align}
& \Pi_{21-pred}^{i, j} = \dfrac{\exp\big(\langle G_2^{i}, G_1^{j}\rangle / \tau\big)}{\sum_{k=1}^{n_1}\exp\big(\langle G_2^{i}, G_1^{k} \rangle / \tau\big)}.\label{eq:diff_p2p}
\end{align}
Here $\langle \cdot,\cdot \rangle$ is the scalar product measuring the similarity between $G_1$ and $G_2$, and $\tau$ is a temperature hyper-parameter. $\Pi_{21-pred}$ can be seen as a soft point-to-point map, formulated based on the adjoint conversion method described in \cite{Pai_2021_CVPR}, and computed in a differentiable manner, hence it can be used inside a neural network.

\begin{figure}
    \centering
    \includegraphics[width=\columnwidth]{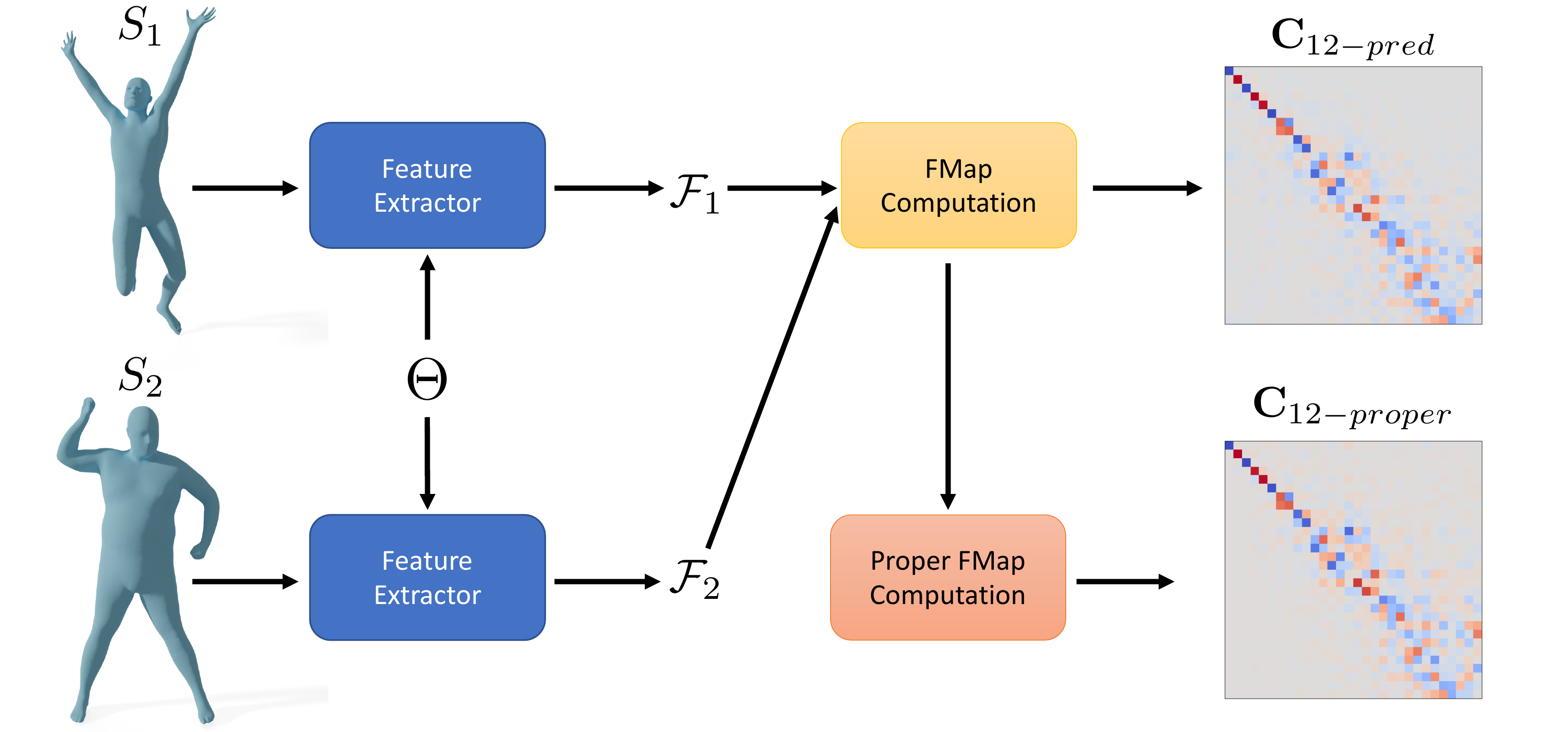}
    \caption{An overview of our revised deep functional map pipeline. The extracted features are used to compute the functional map and the proper functional map, as explained in \cref{sec:proper_fmap}}
    \label{fig:fmap-pipeline}
    \vspace{-1em}
\end{figure}

\paragraph{The feature-based method}
The feature-based method is similar to the adjoint method in spirit, the only difference being that $\Pi_{21-pred}$ is computed using the predicted features instead of the fmap. For this, we use \cref{eq:diff_p2p}, with $G_1 = F_1$ and $G_2 = F_2$. The modified deep functional map pipeline is illustrated in \cref{fig:fmap-pipeline}.

In addition to $C_{12-pred}$, the two previous methods allow to calculate $C_{21-proper}$. We adapt the functional map losses to take into account this modification.

In the supervised case, we modify the supervised loss (see \cref{eq:sup_loss}) by simply introducing an additional term into the loss: 
\begin{align}
\mathcal{L}_{proper} = \| \C_{12-pred} - \C_{12-proper} \|_F^2 \label{eq:sup_loss_proper}.
\end{align}

The motivation behind this loss is that we want the predicted functional map to be as close as possible to the ground truth and stay within the space of proper functional maps.

In the unsupervised setting, we simply impose the standard unsupervised losses on the differentiable proper functional map $\C_{12-proper}$ instead of $\C_{12-pred}$. Specifically, in our experiments below, we use the following unsupervised losses: 
\begin{align}
\nonumber
\mathcal{L}_{unsup}(\C_{12}, \C_{21}) &= \| \C_{12} \C_{21} - \mathbb{I} \|_F^2 + \| \C_{21} \C_{12} - \mathbb{I} \|_F^2 \\
& + \| \C_{12}^{\top} \C_{12} - \mathbb{I} \|_F^2 + \| \C_{21}^{\top} \C_{21} - \mathbb{I} \|_F^2
\label{eq:unsup_loss_proper}
\end{align}




\subsection{As Smooth As Possible Feature Extractor}
Another fundamental assumption of \cref{thm:equivalence} is the completeness of the features produced by a neural network. 

We have experimented with several ways to impose it and have found that it is not easy to satisfy it exactly in general because it would require the network to always produce features in some target  subspace, which is not explicitly specified in advance. Moreover, we have found that explicitly projecting feature functions to a small reduced subspace can also hinder learning. 

To circumvent this, we propose instead to \textit{encourage} this property by promoting the feature extractor to produce smooth features. 

The motivation for this is as follows. If $F_i$ is complete, then there exist coefficients $a_1 ... a_k$ such that $F_i = \sum_{j=1}^k a_j \Phi_i^j$, where $k$ is the size of the functional map used in \cref{eq:fmap_basic}.
However, it's known that Fourier coefficients for smooth functions decay rapidly (faster than any polynomial, if $f$ is of class $C^l$, the coefficients are $o(n^{-l})$), which means that the smoother the function is, the closer it will be to being complete for some index $k$.

\begin{figure}
    \centering
    \includegraphics[width=\columnwidth]{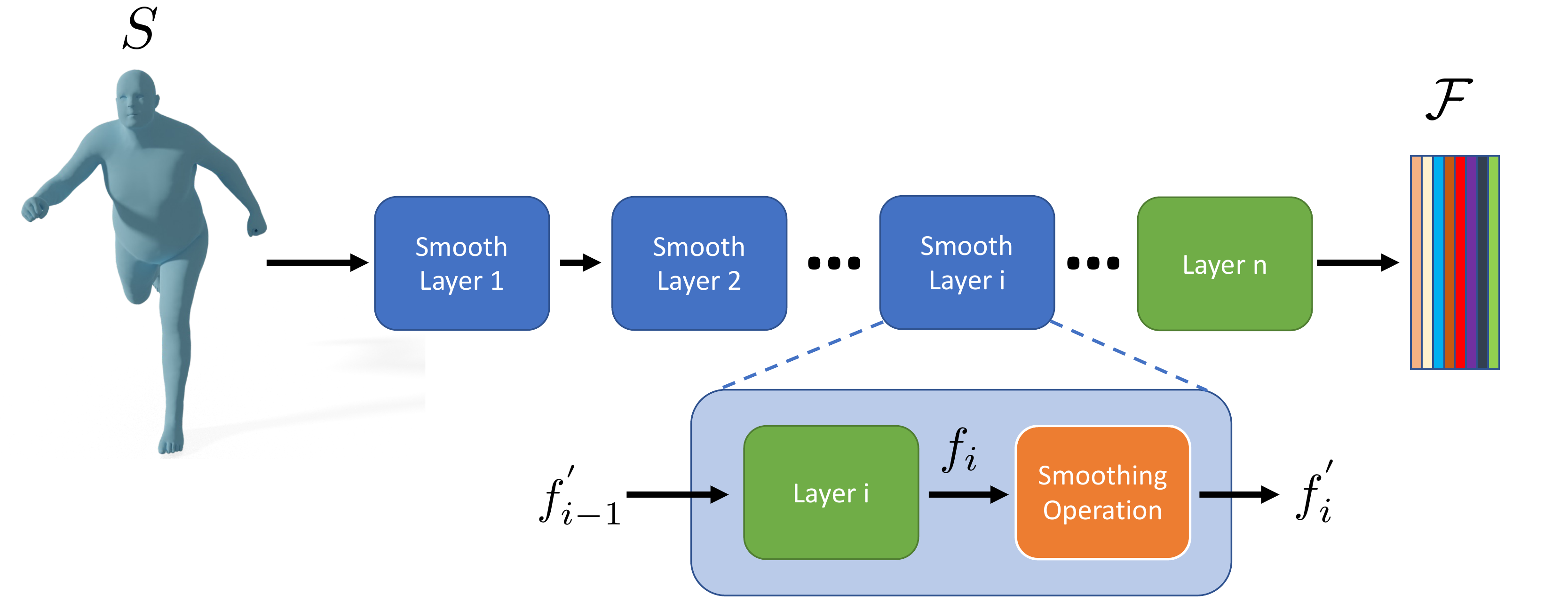}
    \caption{An overview of the feature extractor modification is shown here. The features are made smooth by projecting them into the Laplacian basis at the end of each layer.}
    \label{fig:feat-extract-modif}
\end{figure}

Inspired by this, we propose the following simple modification to feature extractors used for deep functional maps. Since feature extractors are made of multiple layers, we propose to project the output of each layer into the Laplacian basis, diffuse it over the surface following \cite{sharp2021diffusion}, and then project it back to the ambient space before feeding it to the next layer, see \cref{fig:feat-extract-modif}. Concretely, for shape $S$, if $f_i$ is the output of layer $i$, we feed to layer $i+1$ the function $f^{'}_i$, such that $f^{'}_i = \Phi_j e^{-t \Delta} \Phi_j^{\dagger} f_i$, where $\Phi_j$ denotes the first $j$ eigenfunctions of the Laplacian, $\Delta$ is a diagonal matrix containing the first j eigenvalues, and $t$ is a learnable parameter. Please note there is no need to do this operation for the final layer, since the features will be projected into the Laplacian basis anyway, for computing the functional map. In practice, we observed that it is beneficial to set $j$ to \textit{be larger} than the size of the functional maps in \cref{eq:fmap_basic}. This allows the network to impose smoothness, while still allowing degrees of freedom to enable optimization.

%

%
%
%
%
%
%
\vspace{-1em}

\paragraph{Implementation details} we provide implementation details, for all our experiments, in the supplementary. Our code and data will be released after publication.

\section{Results \& Applications}
\label{sec:application}

In this section, we provide results in a wide range of challenging tasks, showing the efficiency and
applicability of our approach to different types of data and feature extractors.

\paragraph{Datasets} 
Our method is evaluated on five datasets commonly used in literature, which include both human and animal datasets.

To evaluate our method's performance for shape matching with humans, we use three datasets: FAUST \cite{Bogo2014}, SCAPE \cite{Anguelov2005}, and SHREC'19 \cite{shrec19}. We utilize the remeshed versions of the first two datasets introduced in \cite{Ren2019} and used in various follow-up works such as \cite{donati2020deep,sharma2020weakly,sharp2021diffusion,eisenberger2020deep,Eisenberger2021NeuroMorphUS,Eisenberger2020SmoothSM}. We follow the same train/test splits as in prior works.
For unsupervised experiments, we utilize the datasets' oriented versions, as described in \cite{sharma2020weakly}, denoted as \textbf{FA}, \textbf{SA}, and \textbf{SHA} for FAUST remeshed aligned, SCAPE remeshed aligned, and SHREC aligned, respectively.

We also evaluate our method on human segmentation using the dataset introduced in \cite{maron2017convolutional}, which comprises segmented human models from various prior datasets. We use the same test split as in prior works.

For animal datasets, we use the SMAL-based dataset \cite{Zuffi:CVPR:2017,marin22_why} (denoted as \textbf{SMAL} hereafter), which consists of 50 organic, non-isometric, and non-rigid shapes represented as 3D meshes. We divide them into 25/25 shapes for training and testing, and to test the generalization capacity of our models, the animals and their positions during testing are never seen during training.

\subsection{Non-Rigid Shape Matching}
In this section, we evaluate our modification to the deep functional map pipeline in three different and challenging settings, including near-isometric supervised matching in \cref{sec:supervised-human}, near-isometric unsupervised shape matching in \cref{sec:unsup-human}, and finally, non-isometric non-rigid matching, both in supervised and unsupervised settings, in \cref{sec:animals-matching}.

We evaluate our proposed modifications in the presence of three different feature extractors, an intrinsic feature extractor DiffusionNet \cite{sharp2021diffusion} which is considered as the state of the art for shape matching, and two extrinsic feature extractors, DGCNN \cite{dgcnn} and DeltaConv \cite{Wiersma2022DeltaConv}, which operate on point clouds. We will show that with our modifications, the latter feature extractors can surpass DiffusionNet in some scenarios.


\paragraph{Evaluation protocol} Our evaluation protocol, following \cite{Kim2011}, measures the geodesic error between predicted maps and the given ground truth, normalized by the square root of the total surface area. All reported values are multiplied by $\times 100$ for clarity.


In our experiments, we use the notation ``X on Y'' to indicate training on X and testing on Y. We use \textbf{FM} to represent the point-to-point map derived from the functional map and \textbf{NN} for the point-to-point map extracted by nearest neighbor in the feature space. We use ``Feature Extractor - Ours'' to denote our training approach for DGCNN and DeltaConv, which involves training the network with both smoothness and properness enforced, while for DiffusionNet, we only enforce properness since smoothness is enforced by construction. We report properness results only for the feature-based method and include adjoint method results in the supplementary materials.


\subsubsection{Supervised Shape Matching}
\label{sec:supervised-human}

\begin{table}[!t]
    \centering
     \ra{1.0}
          \resizebox{\columnwidth}{!}{%
               \begin{tabular}{@{}l cc cc cc cc cc cc@{}}

\rowcolor{Gray!50}
\textbf{Model / Dataset} & \multicolumn{2}{c}{\textbf{FR on FR}} & \multicolumn{2}{c}{\textbf{SR on SR}} & \multicolumn{2}{c}{\textbf{FR on SR}} & \multicolumn{2}{c}{\textbf{SR on FR}} & \multicolumn{2}{c}{\textbf{FR on SH}} & \multicolumn{2}{c}{\textbf{SR on SH}}\\

& FM & NN & FM & NN & FM & NN & FM & NN & FM & NN & FM & NN\\

\midrule
\rowcolor{JungleGreen!80}
DiffusionNet & 2.6 & 2.2 & \textbf{2.9} & 2.7 & \textbf{3.4} & \textbf{3.1} & 2.9 & 3.2 & 9.6 & 7.6 & 6.9 & 9.2 \\
\rowcolor{JungleGreen!80}
DiffusionNet - Ours & \textbf{2.6} & \textbf{2.0} & \textbf{2.9} & \textbf{2.4} & \textbf{3.4} & 3.2 & \textbf{2.7} & \textbf{2.3} & \textbf{5.7} & \textbf{5.7} & \textbf{5.0} & \textbf{5.6} \\

\addlinespace

\rowcolor{Apricot!80}
DGCNN & 2.9 & 6.0 & 3.6 & 8.0 & 13.6 & 16.8 & 3.5 & 10.2 & 11.5 & 17.1 & 10.4 & 19.5 \\
\rowcolor{Apricot!80}
DGCNN - Ours & \textbf{2.6} & 2.6 & 3.1 & 3.3 & 4.1 & 4.8 & 3.4 & 4.6 & 5.8 & 6.0 & 13.8 & 15.2\\

\addlinespace

\rowcolor{Salmon!80}
DeltaConv & 2.7 & 13.8 & 4.4 & 21.2 & 18.1 & 26.6 & 12.2 & 24.4 & 19.4 & 32.6 & 27.7 & 34.9\\
\rowcolor{Salmon!80}
DeltaConv - Ours & \textbf{2.6} & 2.5 & \textbf{2.9} & 2.9 & 3.6 & 4.2 & 5.8 & 6.0 & 10.7 & 10.0 & 22.1 & 21.1\\

\bottomrule

               \end{tabular}
          }
          \caption{Mean geodesic error of various feature extractors for supervised shape matching.}
          \label{tab:supervised_humans}
\end{table}

In this section, we test the quality of our proposed modifications in the supervised setting. For this, we follow the same pipeline described in \cref{sec:background}, and train our network with the supervised loss in \cref{eq:sup_loss_proper}. Also, to stress the generalization power of each network, we do multiple tests, by training on one dataset, and testing on another one, following multiple previous works \cite{donati2020deep,sharma2020weakly,eisenberger2020deep}. For this, we use the \textbf{FR}, \textbf{SR}, and \textbf{SH} datasets. The results are summarized in \cref{tab:supervised_humans}.

First of all, we can see that our approach improves the global result for all feature extractors, on all datasets. The performance increase can be up to 80\%, as is the case for DeltaConv and DGCNN on the scenario \textbf{FR} on \textbf{SR}, where \eg the error has decreased from 18.1 to 3.6.

\begin{figure}
    \centering
    \includegraphics[width=\columnwidth]{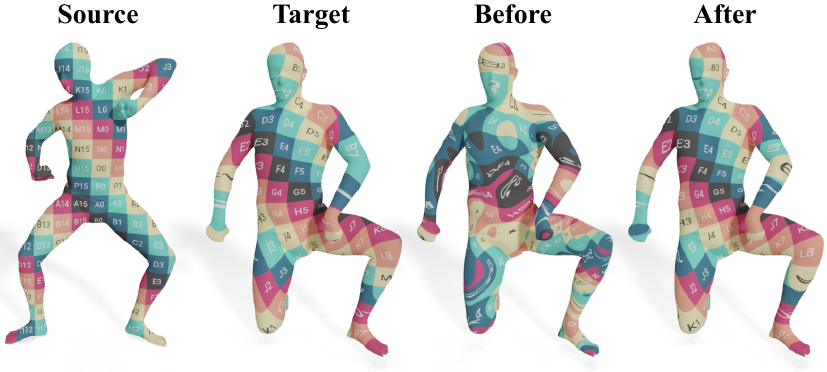}
    \caption{ Qualitative results on SCAPE Remeshed dataset using DeltaConv \cite{Wiersma2022DeltaConv}, before and after our modifications.}
    \label{fig:hum_matching_one}
\end{figure}

Another interesting result to note is that using our modifications, the extrinsic feature extractors, which were failing in the generalization scenarios, now perform as well as DiffusionNet.
Finally, and perhaps more \textit{importantly}, after our modifications, the results obtained with the nearest neighbor are much better, and get closer to the maps extracted by converting the functional map, demonstrating, for the first time, that the features learned in the deep fmap framework do have a geometric meaning, and can be used directly for matching, even without the need to compute a functional map. We explain the slight discrepancy between the result obtained with fmap and NN by the fact that, in practice, the conditions of the theorem are not 100\% satisfied, since we only used approximations to obtain them. In \cref{fig:hum_matching_one}, we show the quality of the produced maps, before and after our modifications. It can be seen that our modifications produce visually plausible correspondences.

\subsubsection{Unsupervised Shape Matching}
\label{sec:unsup-human}

In this section, we test the quality of our modifications in the case of unsupervised learning on near-isometric data. We follow the same method as explained in \cref{sec:background} with the loss in \cref{eq:unsup_loss_proper}, and use three feature extractors: DGCNN, DiffusionNet and DeltaConv, and three datasets: \textbf{FA}, \textbf{SA} and \textbf{SHA}. The results are summarized in \cref{tab:unsup_humans}.


As for the supervised case, our modifications improve the results for all scenarios and for all feature extractors. For example, the result improved by over 70 \% for DeltaConv on the \textbf{FRA} on \textbf{SHA} scenario.

In addition, the extrinsic feature extractors are now on par with DiffusionNet, and the NN results are now close to those obtained with fmap, reinforcing the conclusions obtained in the supervised case \cref{sec:supervised-human}.

\begin{table}[!t]
     \centering
     \ra{1.0}
          \resizebox{\columnwidth}{!}{%
               \begin{tabular}{@{}l cc cc cc cc cc cc@{}}

\rowcolor{Gray!50}
\textbf{Model / Dataset} & \multicolumn{2}{c}{\textbf{FA on FA}} & \multicolumn{2}{c}{\textbf{SA on SA}} & \multicolumn{2}{c}{\textbf{FA on SA}} & \multicolumn{2}{c}{\textbf{SA on FA}} & \multicolumn{2}{c}{\textbf{FA on SHA}} & \multicolumn{2}{c }{\textbf{SA on SHA}}\\

& FM & NN & FM & NN & FM & NN & FM & NN & FM & NN & FM & NN\\

\midrule
\rowcolor{JungleGreen!80}
DiffusionNet & 3.9 & 6.5 & 4.5 & 6.5 & 5.4 & 8.5 & 3.7 & 6.0 & 6.1 & 11.9 & 6.0 & 10.6 \\
\rowcolor{JungleGreen!80}
DiffusionNet - Ours & \textbf{3.3} & \textbf{2.6} & \textbf{3.9} & \textbf{3.4} & \textbf{4.2} & \textbf{4.0} & \textbf{3.3} & \textbf{2.7} & 6.2 & \textbf{5.7} & \textbf{5.3} & \textbf{5.3} \\

\addlinespace

\rowcolor{Apricot!80}
DGCNN & 3.9 & 9.3 & 5.0 & 11.2 & 7.0 & 13.6 & 4.1 & 11.9 & 6.7 & 17.1 & 6.5 & 16.7\\
\rowcolor{Apricot!80}
DGCNN - Ours & 3.9 & 2.8 & 4.6 & 3.8 & 5.2 & 5.0 & 4.0 & 3.4 & 6.5 & \textbf{5.7} & 6.2 & 6.0\\

\addlinespace

\rowcolor{Salmon!80}
DeltaConv & 3.8 & 12.9 & 4.7 & 15.5 & 5.1 & 17.4 & 4.0 & 16.5 & 7.0 & 23.6 & 6.7 & 25.0 \\
\rowcolor{Salmon!80}
DeltaConv - Ours & 3.6 & 3.5 & 4.4 & 4.0 & 4.7 & 4.7 & 4.0 & 3.5 & \textbf{6.0} & 6.1 & 6.7 & 7.7 \\

\bottomrule

               \end{tabular}
          }
          \caption{Mean geodesic error of various feature extractors for unsupervised shape matching.}
          \label{tab:unsup_humans}

     \vspace{-1em}
\end{table}

Note that the effect of enforcing properness is more visible in this context. This can be seen for example in the case of DiffusionNet in Table \cref{sec:supervised-human} since for this feature extractor the only change brought about by our method is via properness. In fact, in the supervised setting, since the training is done using the ground truth-functional maps, and these are proper, this forced the network to learn features that produce functional maps that are as proper as possible, whereas in the unsupervised setting, the network is trained with losses that impose structural properties on the functional maps such as orthogonality, and no properness is involved. We can see that after imposing properness, both functional map and NN results improve, while also getting closer to each other.

\subsubsection{Non-Isometric Shape Matching}
\label{sec:animals-matching}
\begin{table}[!t]
     \centering
     \ra{1.0}
          \resizebox{0.7\columnwidth}{!}{%
               \begin{tabular}{@{}l cc cc@{}}

\rowcolor{Gray!50}
\textbf{Animals Dataset} & \multicolumn{2}{c}{\textbf{Supervised}} & \multicolumn{2}{c}{\textbf{Unsupervised}} \\

& FM & NN & FM & NN \\
\toprule

\rowcolor{JungleGreen!80}
DiffusionNet & 6.1 & 8.7 & 8.0 & 16.6 \\
\rowcolor{JungleGreen!80}
DiffusionNet - Ours & 5.7 & 7.8 & 6.4 & 10.1 \\

\addlinespace

\rowcolor{Apricot!80}
DGCNN & 8.3 & 18.4 & 10.9 & 19.1 \\
\rowcolor{Apricot!80}
DGCNN - Ours & \underline{5.0} & \underline{4.8} & \underline{5.3} & \textbf{5.5} \\

\addlinespace

\rowcolor{Salmon!80}
DeltaConv & 5.4 & 20.7 & 9.5 & 24.1 \\
\rowcolor{Salmon!80}
DeltaConv - Ours & \textbf{4.7} & \textbf{4.2} & \textbf{5.1} & \underline{5.6}\\

\bottomrule

               \end{tabular}
          }
          \caption{Mean geodesic error comparison of different feature extractors on the \textbf{SMAL} dataset, using supervised and unsupervised methods. The highest performing result in each column is denoted in \textbf{bold}, while the second best is denoted with \underline{underline}.}
          \label{tab:match_animals}
      \vspace{-0.5em}
\end{table}

In this section, we test the utility of our modifications in the case of non-rigid non-isometric shape matching, on the \textbf{SMAL} dataset. We test it both in the case of supervised and unsupervised learning, following the same procedure as in \cref{sec:supervised-human} and \cref{sec:unsup-human}. The results are summarized in \cref{tab:match_animals}. As in the previous sections, our modifications improve the results for all feature extractors for both supervised and unsupervised matching. For example, the results for supervised DeltaConv improve by approximately 80\%, from 20.7 to 4.2.

More surprisingly, using our modifications, in the non-isometric setting, DiffusionNet is surpassed as the best feature extractor for shape matching, as DGCNN and DeltaConv achieve better results, using either the functional map or NN method to extract the p2p map. 



\subsection{Generalization Power of Geometric Features}




We conclude our experiments by evaluating the generalization ability of the learned features in the deep functional map pipeline. Our goal is to investigate whether the extracted features contain sufficient information to be used for other tasks without the need for significant fine-tuning.

To conduct this experiment, we employ the human segmentation dataset presented in \cite{maron2017convolutional}. We first extract features from this dataset using a network that has been pre-trained on shape matching. We use the pre-trained DiffusionNet on the \textbf{SA} dataset with the unsupervised. Next, we refine the features for the segmentation task using only a small fraction of the training data. To accomplish this, we utilize a point-wise MLP to avoid complex convolution and pooling operations and to provide an accurate measure of the feature quality.

\begin{figure}
    \centering
    \includegraphics[width=\columnwidth]{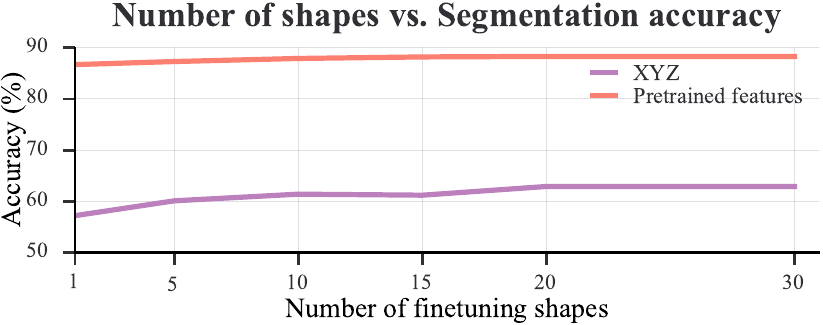}
    \caption{Evolution of human segmentation accuracy as a function of the number of shapes used for the pointwise MLP training.}
    \label{fig:hum_seg_ev}
     \vspace{-0.5em}
\end{figure}

In \cref{fig:hum_seg_ev}, we plot the accuracy of the MLP 
as a function of the number of finetuning shapes. First note that the accuracy of the results improves with the number of training shapes (up to a certain point, where most likely the limited capacity of pointwise MLP is reached). Note also that our features, which are pre-trained in a completely unsupervised manner, produce \textit{significantly} higher accuracy than the raw XYZ coordinates. We attribute this feature utility in a downstream task to the fact that these features capture the geometric structure of the shapes in a compact and invariant manner.
To further demonstrate the generalization power of the pretrained features, we conducted a similar experiment on the RNA dataset \cite{poulenard2019effective} and present the results in the supplementary material.



\tightpara{Ablation Studies} 
In the supplementary, we report an ablation study that demonstrates the effectiveness of the individual components we introduced, in addition to a comparison of our modified pipeline to other recent baseline methods for shape matching.

\section{Conclusion \& Limitations}
\label{sec:conclusion}

In this work, we studied the information contained in the features learned in the deep functional map pipelines, an aspect that has been neglected by all previous works. For this, we established a theorem according to which, under certain mild conditions, these features allow for \textit{direct} computation of point-to-point maps. Inspired by our analysis, we introduced two simple modifications to the deep functional map pipeline. We have shown that applying these modifications improves the overall accuracy of the pipeline in different scenarios, including supervised and unsupervised matching of non-rigid shapes, and makes the features usable for other downstream tasks.

A limitation of our approach is that it is currently geared towards \textit{complete} shape correspondence and must be adapted for partial shape matching. Moreover, it will be interesting to investigate how feature smoothness can be imposed in the context of noisy point clouds or other domains such as graphs or images.

\vspace{-3mm}
\paragraph{Acknowledgements}
The authors acknowledge the anonymous reviewers for their valuable suggestions. 
Parts of this work were supported by the ERC Starting Grant No. 758800 (EXPROTEA) and the ANR AI Chair AIGRETTE.


\newpage
\twocolumn[{%
 \centering
 {\Large \bf Supplementary Materials for:\\Understanding and Improving Features Learned in Deep Functional Maps \par}
 {\vspace*{24pt}}
      {
      \large
      \lineskip .5em
      \begin{tabular}[t]{c}
        Souhaib Attaiki \hspace{3cm}  Maks Ovsjanikov\\
LIX, \'Ecole Polytechnique, IP Paris
      \end{tabular}
      \par
      }
      \vskip .5em
      \vspace*{12pt}
}]

\appendix



In this document, we collect all the results and discussions, which, due to the page limit, could not find space in the main manuscript.

More precisely, we first provide the proof of the theorem we introduced in Sec. 3.3 of the main text in \cref{sec:proof}. Then, the details of the implementation are provided in \cref{sec:implementation}. We present a more in-depth analysis of the modifications we introduced to the deep functional map pipeline in \cref{sec:ablation}. A comparison with other shape-matching methods is provided in \cref{sec:comparison}, as well as additional results regarding the generalization power of pre-trained features in \cref{sec:generalization-power}. Finally, some qualitative results are included in \cref{sec:qualitative}.




\section{Proof of Theorem 3.1}
\label{sec:proof}

In Sec. 3.3 of the main text, we stated a theorem that shows that the maps obtained with the adjoint method, or the nearest neighbor in the feature space are equivalent under some conditions. In this section, we formally restate it and provide proof.

As mentioned in the main body, in our result below we assume that all optimization problems have unique global minima. Thus, for the problem $\argmin_{\C} \|\C \A_{1} - \A_{2} \|$, this means that $\A_1$ must be full row rank, whereas, for the problem of type $\argmin_{\Pi} \| \Pi F_1 - F_2 \|$, this means that the rows of $F_1$ must be distinct (i.e., no two rows are identical, as vectors).

\begin{customthm}{3.1}
\label{thm:equivalence_supp}
    Suppose the feature extractor $F_{\Theta}$ is \emph{complete}. Let $\A_1 = \Phi_1^{+} F_1$ and $\A_2 = \Phi_2^{+} F_2$. Then, denoting $\C_{\rm{opt}} = \argmin_{\C} \|\C \A_{1} - \A_{2} \|$, we have the following results hold:
    
(1) If $\Pi F_1 = F_2$ for some point-to-point map $\Pi$ then $\C_{12} = \Phi_2^{+} \Pi \Phi_1$ is basis-aligning. Moreover, $\C_{12} = \C_{\opt}$ and extracting the pointwise map from $\C_{\opt}$ via the adjoint method, or via nearest neighbor search in the feature space $\min_{\Pi} \| \Pi F_1 - F_2 \|$ will give the same result.

(2) Conversely, suppose that $\C_{\opt}$ is basis aligning, then $\argmin_{\Pi} \| \Pi F_1 - F_2 \| = \argmin_{\Pi} \| \Pi \Phi_1 - \Phi_2 C_{\opt}\|$.
\end{customthm}

\begin{proof}

(1) If $\Pi F_1 = F_2$ and $F_1, F_2$ are complete by assumption then we have $F_1 = \Phi_1 \A_1$ and $F_2 = \Phi_2 \A_2$ so that:
\begin{equation}
\label{eq:p1}
    \Pi \Phi_1 \A_1 = \Phi_2 \A_2
\end{equation}

Setting $\C_{12} = \Phi_2^{+} \Pi \Phi_1$ and pre-multiplying Eq.~(\ref{eq:p1}) by $\Phi_2^{+}$ we obtain $\C_{12} \A_1 = \A_2.$ Thus, $\|\C_{12} \A_1 - \A_2\| = 0$, and it follows that $\C_{\opt} = \C_{12}$, since $\A_1$ assumed to be of full row rank (and thus $\argmin_{\C} \|\C \A_{1} - \A_{2} \|$ has a unique  optimum).

Moreover, using $\C_{12} \A_1 = \A_2,$ we get $\Phi_2 \A_2 = \Phi_2 \C_{12} \A_1$. Combining this with \cref{eq:p1}, we get $\Pi \Phi_1 \A_1 = \Phi_2 \C_{12} \A_1$. Using the fact that $\A_1$ is full rank, this implies that $\Pi \Phi_1 = \Phi_2 \C_{12}$, and thus $\C{12}$ is basis-aligning.

Finally, we note that since the same pointwise map satisfies $\|\Pi F_1 - F_2 \| = \|\Pi \Phi_1 - \Phi_2 \C_{12}\| = 0$, minimizing both energies with respect to $\Pi$ would result in the same map. 
	 
(2) By assumption $\C_{\opt}$ is basis-aligning. Thus, $\Pi_{21} \Phi_1 = \Phi_2 \C_{\opt}$ for some pointwise map $\Pi_{21}$. Thus,

\begin{equation}
\label{eq:p2}
    \min_{\Pi} \| \Pi \Phi_1 - \Phi_2 \C_{\opt}\| = \Pi_{21}
\end{equation}

Now let's consider the problem 
\begin{equation}
\label{eq:p3}
    \min_{\Pi} \|\Pi F_1 - F_2\| = \min_{\Pi} \|\Pi F_1 - F_2\|^2
\end{equation}

The objective can be decomposed into two parts, one that lies within the span of $\Phi_2$ and the one outside of it, as follows: 
\begin{align*}
   E(\Pi) =&  \|\Pi F_1 - F_2\|^2 \nonumber \\
   =&\|\Phi_2^{+}(\Pi F_1 - F_2)\|^2 + \|(I - \Phi_2 \Phi_2^{+})(\Pi F_1 - F_2)\|^2  \nonumber \\
   =& E_1(\Pi) + E_2(\Pi) \nonumber 
\end{align*}
 
Using the fact that $F_i$ are complete, we have $F_i = \Phi_i \A_i$. It follows that $E_1 = \|\Phi_2^{+} \Pi \Phi_1 \A_1 - \A_2\|^2$.

On the other hand, using the fact that $F_2$ is complete,  we have $(I - \Phi_2 \Phi_2^{+})F_2 = 0 $, and thus $E_2 = \|(I - \Phi_2 \Phi_2^{+})\Pi F_1 \|^2$. 

Now recall that by assumption $\C_{\opt}$ is basis-aligning and thus there exists some pointwise map $\Pi_{21}$ such that $\Pi_{21} \Phi_1 = \Phi_2 \C_{\opt}$.

Consider $E_1(\Pi)$ for an arbitrary pointwise map $\Pi$. We have:
\begin{align*}
     \min_{\C} \|\C \A_{1} - \A_{2} \|         =& \|C_{\opt} \A_1 - \A_2\| \\
              =& \|\Phi_2^{+} \Pi_{21} \Phi_1 \A_1 - \A_2\| \\
              =& E_1(\Pi_{21})
\end{align*}

It follows that we must have that $E_1(\Pi_{21}) \le E_1(\Pi)$ for any pointwise map $\Pi$.

Moreover observe that for $\Pi_{21}$ we have:
\begin{align*}
   E_2(\Pi_{21}) =&  \|(I - \Phi_2 \Phi_2^{+})\Pi_{21} F_1 \|^2 \nonumber \\
   =&\|(I - \Phi_2^{+} \Phi_2^{+})\Pi_{21} \Phi_1 \A_1 \|^2  \nonumber \\
   =& \|(I - \Phi_2 \Phi_2^{+}) \Phi_2 \C_{\opt} \A_1 \|^2 \nonumber \\
   =& \| \Phi_2 \C_{\opt} \A_1 - \Phi_2 \C_{\opt} \A_1\| \nonumber \\
   =& 0
\end{align*}

Thus, for an arbitrary pointwise map $\Pi$ we must have $E_2(\Pi_{12}) \le  E_2(\Pi)$. 

It therefore follows that $\argmin_{\Pi} \|\Pi F_1 - F_2\| = \argmin_{\Pi} \left(E_1(\Pi) + E_2(\Pi)\right) = \Pi_{21}$, and thus $\argmin_{\Pi} \| \Pi F_1 - F_2 \| = \argmin_{\Pi} \| \Pi \Phi_1 - \Phi_2 C_{\opt}\|.$
\end{proof}

\paragraph{Verification of the assumptions of the theorem} Note that as mentioned above, we first assumed that $\A_1$ is of full rank, and second, the rows of $F_1$ must be distinct so that all optimization problems have unique minima. We would like to point out that these assumptions are very weak and easily hold in practice.

Indeed, as far as the second condition is concerned, it is easily verifiable because of numerical precision, as it is very unlikely that two distinct points are assigned the exact same feature vectors. In practice, for example, we compute for each point in a feature produced by DiffusionNet, the distance to its nearest neighbor, and we take the average over the whole shape. We find that this distance is equal to 0.15, while for reference, it is equal to 0.0004 for the XYZ coordinates, which shows that this assumption is justified. 

\begin{figure}
    \centering
    \includegraphics[width=\columnwidth]{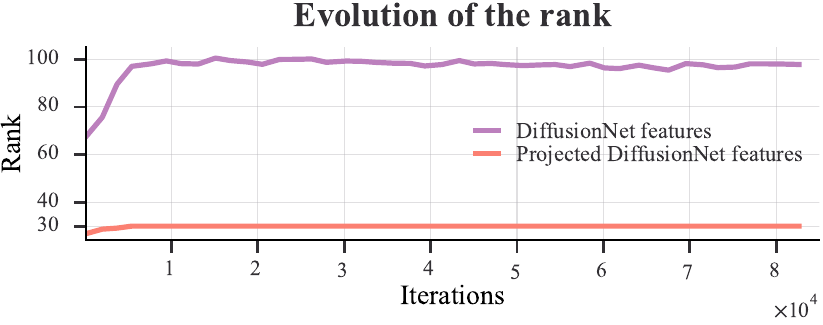}
    \caption{Evolution of the rank of features produced by DiffusionNet, as well as their projection in the Laplacian-Basis, during training.}
    \label{fig:rank_evolution}
\end{figure}

For the first assumption, we plot in \cref{fig:rank_evolution} the evolution of the rank of the features produced by DiffusionNet during training, as well as the evolution of the rank of their projection on the spectral basis (\ie $\A_i$). We can see that throughout the training, the rank of the projected features is always equal to 30, which is the same dimension as the spectral basis used in all our experiments, which proves that the assumption of the full rank of $\A_1$ is verified.

\paragraph{Impact of the proposed modifications on the conditions of the theorem}
In this paragraph, we provide measures of the individual conditions of the theorem before and after applying our modifications in order to shed light on their effectiveness.

Given a functional map $\C_{12}$ predicted by the fmap framework, we compute its basis aligning property by computing the Chamfer distance between $\Phi_2 \C_{12}$ and $\Phi_1$ using the notation in Definition 3.2 of the main text (note that $\|\Phi_2 \C_{12} - \Pi_{21} \Phi_1\|$ is a one-way Chamfer distance). 
For properness, given a predicted functional map $\C$, we compute its proper version $\C_\mathrm{proper}$ using the adjoint method and measure properness by computing $\Vert \C - \C_\mathrm{proper}\Vert_2^2$. 
For completeness, using the same notation as in Eq. (4), we measure it using $1 - \frac{\Vert \mathcal{F}_\Theta (S_i) - \Phi_i \Phi_i^{\dagger} \mathcal{F}_\Theta (S_i)\Vert_2^2}{\Vert \mathcal{F}_\Theta (S_i) \Vert_2^2 }$, this quantity is between 0 and 1 (higher is better). 

We report these measures in \cref{tab:complete_proper} for the same setup used in our ablation study, i.e., unsupervised shape matching on "FA on FA" (see \cref{sec:ablation}), for two main backbones: DGCNN and DeltaConv (note that in DiffusionNet completeness is enforced by construction). It can be seen that our modifications improve the measured properties.

\begin{table}[t]  
    \centering
    \ra{1.0}
         \resizebox{0.95\columnwidth}{!}{
            \begin{tabular}{@{}lcccc@{}}
\toprule
& \multicolumn{2}{c}{DGCNN} & \multicolumn{2}{c}{DeltaConv} \\
Property & before & after & before & after \\
\midrule
Basis alignment ($\downarrow$) & 9.7 & \textbf{4.8} & 11.3 & \textbf{5.9}  \\ 
Properness ($\downarrow$) & 9.6 & \textbf{4.4} & 13.1 & \textbf{7.9}  \\
Completeness ($\uparrow$) & 69 \% & \textbf{90 \%} & 64 \% & \textbf{83 \%} \\
\bottomrule
            \end{tabular}
         }
     \vspace{-1mm}
\caption{Impact of our modifications on the key metrics.}
\label{tab:complete_proper}
\end{table}




\section{Implementation Details}
\label{sec:implementation}
In all our experiments with functional maps, we used the functional map size of $k=30$. For the Laplace-Beltrami computation, we used the discretization introduced in \cite{sharp2020laplacian}.

In our experiments in Sec. 5.2 of the main text, we used three different networks, DGCNN \cite{dgcnn}, DiffusionNet \cite{sharp2021diffusion} and DeltaConv \cite{Wiersma2022DeltaConv}. For this, we used
the publicly available implementations \footnote{\url{https://github.com/WangYueFt/dgcnn}} \footnote{\url{https://github.com/nmwsharp/diffusion-net}} \footnote{\url{https://github.com/rubenwiersma/deltaconv}} released by the authors. In all our experiments, we used the default segmentation configuration provided by the authors in their respective papers, with an output dimension of 128. For all these experiments, we used the Adam optimizer \cite{kingma2017adam}, with a learning rate of 0.001. We used data augmentation in all our experiments.  In particular, we augment the training data on the fly by randomly rotating the input shapes, varying the position of each point by Gaussian noise, and applying random scaling in the interval [0.9, 1.1].

For the proper map computation (Eq.~6 of the main text), we used $\tau = 0.07$.

To make the feature extractor smooth, 
we project the features onto a Laplace-Beltrami basis of size $j=128$, which is the same size used in the original DiffusionNet implementation.

Regarding the experiment in Sec. 5.3 of the main text, we used the pre-trained DiffusionNet model on the Scape Remeshed Aligned dataset, with unsupervised and properness losses. The downstream point MLP consists of 4 layers, and we train it with the Adam optimizer for 300 iterations, using a learning rate of 0.001. The training shape is randomly sampled from the training data set. 


\section{Ablation Study}
\label{sec:ablation}

In Sec. 4 of the main text, we introduced two modifications to the functional map pipeline, namely imposing properness on the functional map and requiring the features to be as smooth as possible. Here we show the effect of each modification independently. For illustration, we will use the unsupervised near-isometric matching experiment (Sec. 5.2.2 of the main text), but a similar conclusion can be drawn in all other scenarios.

The results are summarized in \cref{tab:ablation}. In this table, we show the result for each architecture without modification \textcircled{0}, with properness with the adjoint method \textcircled{1}, with feature-based properness \textcircled{2}, using the smoothness operation \textcircled{3}
or the combination of the latter. Since in DiffusionNet, the smoothing operation is performed by construction, we only include the results with properness.

We can see that each of our modifications improves the result of the vanilla feature extractor, and for optimal performance, both modifications should be used. In particular, we noticed that imposing the properness using the feature-based method gives a slightly better result, so we advocate using this method.

\begin{table}[!t]
    \centering
    \ra{1.0}
          \resizebox{\columnwidth}{!}{%
               \begin{tabular}{@{}L CC CC CC CC CC CC@{}}

\textbf{Model / Dataset} & \multicolumn{2}{c}{\textbf{FA on FA}} & \multicolumn{2}{c}{\textbf{SA on SA}} & \multicolumn{2}{c}{\textbf{FA on SA}} & \multicolumn{2}{c}{\textbf{SA on FA}} & \multicolumn{2}{c}{\textbf{FA on SHA}} & \multicolumn{2}{c }{\textbf{SA on SHA}}\\

& FM & NN & FM & NN & FM & NN & FM & NN & FM & NN & FM & NN\\

\midrule
\rowcolor{JungleGreen!80}[][]
DiffusionNet - \textcircled{0}
& 3.9 & 6.5 & 4.5 & 6.5 & 5.4 & 8.5 & 3.7 & 6.0 & 6.1 & 11.9 & 6.0 & 10.6 \\

\rowcolor{JungleGreen!80}[][]
DiffusionNet  - \textcircled{1}
& 3.4 & 3.2 & 4.2 & 4.4 & 4.3 & 4.9 & 3.4 & 3.7 & \textbf{5.5} & 5.8 & \textbf{5.3} & 6.1 \\

\rowcolor{JungleGreen!80}[][]
DiffusionNet  - \textcircled{2}
& \textbf{3.3} & \textbf{2.6} & \textbf{3.9} & \textbf{3.4} & \textbf{4.2} & \textbf{4.0} & \textbf{3.3} & \textbf{2.7} & 6.2 & \textbf{5.7} & \textbf{5.3} & \textbf{5.3} \\

\addlinespace

\rowcolor{Apricot!80}[][]
DGCNN - \textcircled{0}
& 3.9 & 9.3 & 5.0 & 11.2 & 7.0 & 13.6 & 4.1 & 11.9 & 6.7 & 17.1 & 6.5 & 16.7\\

\rowcolor{Apricot!80}[][]
DGCNN - \textcircled{1}
& \textbf{3.5} & 5.6 & 4.4 & 7.3 & 7.1 & 11.5 & 3.6 & 8.8 & 7.0 & 13.2 & 5.7 & 13.0\\

\rowcolor{Apricot!80}[][]
DGCNN - \textcircled{2}
& 3.7 & 3.5 & 4.8 & 5.2 & 7.4 & 11.3 & 4.0 & 5.1 & 8.0 & 10.7 & 6.3 & 8.4\\

\rowcolor{Apricot!80}[][]
DGCNN - \textcircled{3}
& 3.9 & 4.7 & 4.8 & 5.7 & 5.0 & 6.2 & 3.9 & 5.2 & 6.5 & 7.4 & 6.4 & 8.4\\

\rowcolor{Apricot!80}[][]
DGCNN - \textcircled{1} + \textcircled{3}
& \textbf{3.5} & 3.6 & \textbf{4.3} & 4.4 & \textbf{4.5} & 5.2 & \textbf{3.4} & 4.5 & \textbf{5.4} & 5.8 & \textbf{5.3} & 6.5\\

\rowcolor{Apricot!80}[][]
DGCNN - \textcircled{2} +  \textcircled{3}
& 3.9 & \textbf{2.8} & 4.6 & \textbf{3.8} & 5.2 & \textbf{5.0} & 4.0 & \textbf{3.4} & 6.5 & \textbf{5.7} & 6.2 & \textbf{6.0}\\

\addlinespace

\rowcolor{Salmon!80}[][]
DeltaConv - \textcircled{0}
& 3.8 & 12.9 & 4.7 & 15.5 & 5.1 & 17.4 & 4.0 & 16.5 & 7.0 & 23.6 & 6.7 & 25.0 \\

\rowcolor{Salmon!80}[][]
DeltaConv - \textcircled{1}
& \textbf{3.4} & 7.0 & 4.2 & 9.5 & 4.5 & 13.4 & 3.6 & 12.4 & 5.8 & 18.0 & 6.2 & 19.2 \\

\rowcolor{Salmon!80}[][]
DeltaConv - \textcircled{2}
& 3.6 & \textbf{3.5} & 4.3 & 5.1 & 5.9 & 8.7 & 3.9 & 7.0 & 7.0 & 10.6 & 6.2 & 11.5 \\

\rowcolor{Salmon!80}[][]
DeltaConv - \textcircled{3}
& 3.8 & 5.7 & 4.5 & 6.7 & 4.7 & 7.1 & 3.8 & 5.8 & 6.3 & 9.1 & 6.6 & 11.7 \\

\rowcolor{Salmon!80}[][]
DeltaConv - \textcircled{1} + \textcircled{3}
& \textbf{3.4} & 3.6 & \textbf{4.1} & 4.6 & \textbf{4.3} & 5.2 & \textbf{3.4} & 4.2 & \textbf{5.3} & 6.3 & \textbf{5.4} & \textbf{7.1} \\

\rowcolor{Salmon!80}[][]
DeltaConv - \textcircled{2} + \textcircled{3} 
& 3.6 & \textbf{3.5} & 4.4 & \textbf{4.0} & 4.7 & \textbf{4.7} & 4.0 & \textbf{3.5} & 6.0 & \textbf{6.1} & 6.7 & 7.7 \\

\bottomrule
               \end{tabular}
          }
          \caption{Ablation Study on the components of our method. \textcircled{0}: no modification. \textcircled{1}: use properness based on the adjoint method. \textcircled{2}: use properness using the feature-based method. \textcircled{3}: make the feature extractor as smooth as possible using the method introduced in Sec. 4.2 of the main text. We highlight the best result of each feature extractor in \textbf{bold}.}
          \label{tab:ablation}
\end{table}

As we explained in the main text, the difference between the result obtained with the functional map, and the result with the nearest neighbor (NN) method is explained by the fact that the conditions of the theorem are not fully satisfied. 
Indeed, while the results of the two approaches are generally very close, after our modifications, there is variability in terms of which method produces the best results depending on the dataset. Remark that if the NN approach is better, this suggests that higher frequencies in the learned feature functions are beneficial for correspondence. However, the opposite can (and indeed does) occur, in that those high frequencies can hinder results since they are not penalized during training (as the functional map losses are computed after projecting the features onto a low-frequency basis).

\section{Comparison with other methods}
\label{sec:comparison}
In this section, we compare our method to other recent shape-matching methods and evaluate the effect of our proposed modifications on additional baseline approaches. We compare our method in supervised and unsupervised near-isometric shape-matching experiments in Sec.~5.2.1 and Sec.~5.2.2 of the main text, respectively.

Note that since our proposed modifications are general and can be applied to \textit{any deep functional map pipeline} in principle, we also tested their effects on additional methods. In addition to the comparisons shown in the main manuscript, 
below we also test our modifications on two very recent approaches: SRFeat \cite{li2022srfeat} for the supervised case, and DUO-FMap \cite{donati-duo} for the unsupervised case. Because both SRFeat and DUO-FMap use DiffusionNet as a backbone, we only enforce the properness using the feature-based method, following Sec. 4.1 of the main text.

We use the same protocol as in the main text. Namely, the geodesic error is normalized by the square root of the total surface area, values are multiplied by $\times 100$ for clarity, and the notation ``X on Y'' means that we train on X and test on Y. We denote the methods using our proposed modifications by ``Method - Ours''.

\begin{table}[!t]
    \centering
    \ra{1.0}
          \resizebox{\columnwidth}{!}{%
               \begin{tabular}{@{}LRRRRRR@{}}

\rowcolor{Gray!50}[][]
\textbf{Model / Dataset} & \textbf{FR on FR} & \textbf{SR on SR} & \textbf{FR on SR} & \textbf{SR on FR} & \textbf{FR on SH} & \textbf{SR on SH}\\

\midrule
3D-CODED & 2.5 & 31.0 & 31.0 & 33.0 & -- & -- \\
FMNet & 11.0 & 30.0 & 30.0 & 33.0 & -- & -- \\
HSN & 3.3 & 3.5 & 25.4 & 16.7 & -- & -- \\
GeomFmaps & 3.1 & 4.4 & 11.0 & 6.0 & 9.9 & 12.2 \\
TransMatch & 2.7 & 18.3 & 33.6 & 18.6 & 21.0 & 38.8 \\
GeomFmaps + DiffusionNet & 2.6 & 2.9 & 3.4 & 2.9 & 9.6 & 6.9\\
SRFeat & \textbf{1.1} & \underline{2.2} & 3.9 & 2.5 & 9.9 & 6.2\\

\addlinespace

\rowcolor{RoyalBlue!60}[][]
GeomFmaps + DiffusionNet - Ours & 2.0 & 2.4 &  \underline{3.2}  & \underline{2.3} & \textbf{5.7} & \underline{5.6} \\
\rowcolor{RoyalBlue!60}[][]
SRFeat - Ours & \underline{1.3} & \textbf{1.8} & \textbf{2.9} & \textbf{1.8} & \underline{5.8} & \textbf{5.4}\\

\bottomrule
               \end{tabular}
          }
          \caption{Accuracy of various supervised shape matching methods for near-isometric shape matching. Our modifications achieve state-of-the-art results. The best result in each column is highlighted in \textbf{bold}, and the second best is highlighted using \underline{underline}.}
          \label{tab:comp_supervised}

\end{table}

Concerning the supervised near-isometric shape-matching experiment, we compare to FMNet \cite{litany2017deep}, 3DCODED \cite{groueix20183d}, HSN \cite{wiersma2020cnns}, TransMatch \cite{trappolini2021shape}, GeomFMaps \cite{donati2020deep}, and SRFeat \cite{li2022srfeat}. Results are summarized in \cref{tab:comp_supervised}. It can be seen that our method achieves state-of-the-art results among supervised methods, especially in challenging cases such as testing on the SHREC Remeshed dataset.

For the unsupervised setting, we test our method against SURFMNet \cite{roufosse2019unsupervised}, UnsupFMNet \cite{halimi2019unsupervised}, WSupFMNet \cite{sharma2020weakly}, Deep Shells \cite{eisenberger2020deep}, Neuromorph \cite{Eisenberger2021NeuroMorphUS}, and DUO-FMap \cite{donati-duo}. Results are summarized in \cref{tab:comp_unsupervised}. As can be seen, our method achieves state-of-the-art results in this scenario also, especially in challenging cases involving generalization, where all competing methods fail. It can also be seen that the modifications we propose are complementary to the different versions of the deep functional map pipeline. Remarkably, our method brings consistent and significant improvements throughout \textit{all cases} and baseline approaches that we tested.

We would like to emphasize that our main contribution is both an analysis and a set of \textit{improvements} for the deep functional map pipeline in general. As such, rather than a particularly new approach for correspondence, our key contribution is a set of  modifications, which can be adapted within different deep functional map pipelines. We emphasize this because our approach is flexible and, as illustrated in the results, can be beneficial for different methods proposed in the literature for both supervised and unsupervised cases.

\begin{table}[!t]
    \centering
    \ra{1.0}
          \resizebox{\columnwidth}{!}{%
               \begin{tabular}{@{}LRRRRRR@{}}

\rowcolor{Gray!50}[][]
\textbf{Model / Dataset} & \textbf{FA on FA} & \textbf{SA on SA} & \textbf{FA on SA} & \textbf{SA on FA} & \textbf{FA on SHA} & \textbf{SA on SHA}\\

\midrule
SURFMNet & 15.0 & 12.0 & 30.0 & 30.0 & -- & -- \\
UnsupFMNet & 10 & 16.0 & 29.0 & 22.0 & -- & -- \\
WSupFMNet & 3.3  & 7.3 & 11.7 &  6.2 & -- & -- \\
DeepShells & \textbf{1.7} & \underline{2.5} & 5.4 & 2.7 & 26.3 & 22.8 \\
NeuroMorph & 8.5 & 29.9 & 28.5 & 18.2 & 26.3 & 27.6 \\
WSupFMNet + DiffusionNet & 3.9 & 4.5 & 5.4 & 3.7 & 6.1 & \underline{6.0}\\
DUO-FMap & 2.5 & 2.6 & 4.2 & \underline{2.7} & 6.4 & 8.4 \\
\addlinespace
\rowcolor{RoyalBlue!60}[][]
WSupFMNet + DiffusionNet - Ours & 2.6 & 3.4 & \underline{4.0} & \underline{2.7} & \underline{5.7} & \textbf{5.3} \\
\rowcolor{RoyalBlue!60}[][]
DUO-FMap - Ours & \underline{2.3} & \textbf{2.4} & \textbf{3.0} & \textbf{2.4} & \textbf{5.5} & \textbf{5.3} \\

\bottomrule
               \end{tabular}
          }
          \caption{Accuracy of various unsupervised shape matching methods for near-isometric shape matching. Our modifications achieve state-of-the-art results. The best result in each column is highlighted in \textbf{bold}, and the second best is highlighted using \underline{underline}.}
          \label{tab:comp_unsupervised}

\end{table}



\section{Generalization Power of Pre-Trained Features}
\label{sec:generalization-power}

In Sec. 5.3 of the main text, we showed the generalization power of our pretrained features on the task of human segmentation. Here we show more results consolidating the fact that learned features do have a geometric signification.




For this, we tested the utility of the pre-trained features for the task of molecular surface RNA segmentation. We used the RNA dataset introduced in \cite{poulenard2019effective}, composed of 640 RNA triangle meshes, where each vertex is labeled into one of 259 atomic categories. We used the same 80/20\% split for training and test sets as in previous works.

We follow the same setup as in Sec. 5.3 of the main text. Specifically, we train a DiffusionNet network for shape matching in an unsupervised manner on the RNA data. We then extract features for each shape using the trained network. Finally, these features are used to train another DiffusionNet for the semantic segmentation task in a supervised manner.

The results are summarized in \cref{tab:rna_seg}. We can see that our pre-trained features outperform XYZ and HKS \cite{sun2009concise} (note that due to the relatively large size of the training set, the improvement starts to saturate). We take this as further evidence that the features extracted using our approach encode geometric information that can be useful in various shape analysis tasks.

\begin{table}[!t]
    \centering
    \ra{1.0}
               \begin{tabular}{@{}lr@{}}
\toprule
\textbf{Method} & \textbf{Accuracy} \\
\midrule
SplineCNN \cite{fey2018splinecnn} & 53.6 \% \\
SPHNet \cite{poulenard2019effective} & 80.2 \% \\
SurfaceNetworks \cite{kostrikov2018surface} & 88.5\% \\
\addlinespace
DiffusionNet - XYZ & 90.5\% \\
DiffusionNet - HKS & 90.6\% \\
DiffusionNet - Pretrained features & \textbf{90.8\%} \\
\bottomrule

               \end{tabular}
          \caption{Accuracy of various methods for RNA segmentation.}
          \label{tab:rna_seg}
          
\end{table}

\section{Qualitative Results}
\label{sec:qualitative}

In this section, we present some qualitative results with our method.

\begin{figure}
    \centering
    \includegraphics[width=\columnwidth]{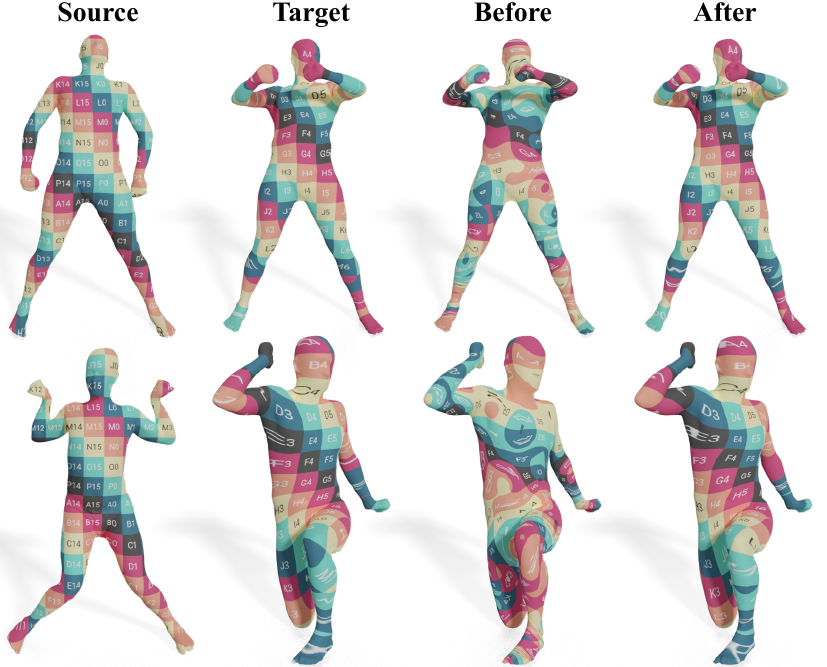}
    \caption{ Qualitative results on SCAPE Remeshed dataset using DeltaConv, before and after using the modifications we have introduced.}
    \label{fig:hum_matching}
\end{figure}

In \cref{fig:hum_matching}, we show the quality of the produced maps, before and after our modifications. It can be seen that our modifications produce visually plausible correspondences.

\begin{figure}
    \centering
    \includegraphics[width=\columnwidth]{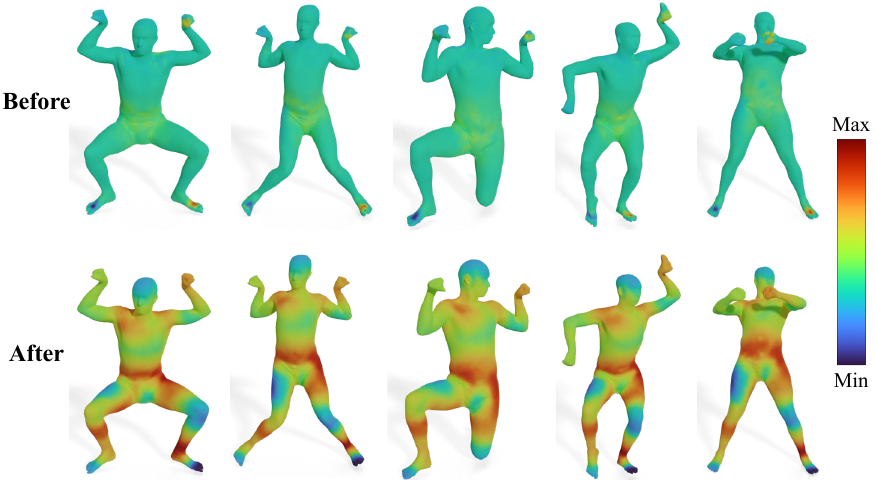}
    \caption{Visualization of the intensity of the same feature learned by DeltaConv for shape matching, on several shapes, before and after using our modifications. The intensity is color coded.}
    \label{fig:feat_rep}
\end{figure}

In \cref{fig:feat_rep}, we show how repeatable the features are, before and after our modifications, the intensity is color coded. We can see that before the modifications, the features are mostly flat and non-distinctive, making them not useful for matching using the nearest neighbor method, or for use in a downstream task, whereas, after our modifications, we can see that the features are activated over the same region, over multiple shapes, and they are not flat since they vary according to the geometry.


{\small
\bibliographystyle{ieee_fullname}
\bibliography{references}

\begin{thebibliography}{10}\itemsep=-1pt

\bibitem{Anguelov2005}
Dragomir Anguelov, Praveen Srinivasan, Daphne Koller, Sebastian Thrun, Jim
  Rodgers, and James Davis.
\newblock {SCAPE}.
\newblock {\em {ACM} Transactions on Graphics}, 24(3):408--416, July 2005.

\bibitem{attaiki2023vader}
Souhaib Attaiki, Lei Li, and Maks Ovsjanikov.
\newblock Generalizable local feature pre-training for deformable shape
  analysis.
\newblock In {\em The IEEE Conference on Computer Vision and Pattern
  Recognition (CVPR)}, June 2023.

\bibitem{attaiki2022ncp}
Souhaib Attaiki and Maks Ovsjanikov.
\newblock {NCP}: Neural correspondence prior for effective unsupervised shape
  matching.
\newblock In {\em Advances in Neural Information Processing Systems}, 2022.

\bibitem{attaiki2021dpfm}
Souhaib Attaiki, Gautam Pai, and Maks Ovsjanikov.
\newblock {DPFM}: Deep partial functional maps.
\newblock In {\em 2021 International Conference on 3D Vision (3DV)}. {IEEE},
  Dec. 2021.

\bibitem{pcnn_2018}
Matan Atzmon, Haggai Maron, and Yaron Lipman.
\newblock Point convolutional neural networks by extension operators.
\newblock {\em ACM Transactions on Graphics}, 37(4):1–12, Aug 2018.

\bibitem{aubry2011wave}
Mathieu Aubry, Ulrich Schlickewei, and Daniel Cremers.
\newblock The wave kernel signature: A quantum mechanical approach to shape
  analysis.
\newblock In {\em 2011 IEEE international conference on computer vision
  workshops (ICCV workshops)}, pages 1626--1633. IEEE, 2011.

\bibitem{bai2021pointdsc}
Xuyang Bai, Zixin Luo, Lei Zhou, Hongkai Chen, Lei Li, Zeyu Hu, Hongbo Fu, and
  Chiew-Lan Tai.
\newblock {PointDSC}: Robust point cloud registration using deep spatial
  consistency.
\newblock In {\em CVPR}, 2021.

\bibitem{Baran2009}
Ilya Baran, Daniel Vlasic, Eitan Grinspun, and Jovan Popovi{\'{c}}.
\newblock Semantic deformation transfer.
\newblock {\em {ACM} Transactions on Graphics}, 28(3):1--6, July 2009.

\bibitem{Bogo2014}
Federica Bogo, Javier Romero, Matthew Loper, and Michael~J. Black.
\newblock {FAUST}: Dataset and evaluation for 3d mesh registration.
\newblock In {\em 2014 {IEEE} Conference on Computer Vision and Pattern
  Recognition}. {IEEE}, June 2014.

\bibitem{bronstein2017geometric}
Michael~M Bronstein, Joan Bruna, Yann LeCun, Arthur Szlam, and Pierre
  Vandergheynst.
\newblock Geometric deep learning: going beyond euclidean data.
\newblock {\em IEEE Signal Processing Magazine}, 34(4):18--42, 2017.

\bibitem{burghard2017embedding}
Oliver Burghard, Alexander Dieckmann, and Reinhard Klein.
\newblock Embedding shapes with {G}reen's functions for global shape matching.
\newblock {\em Computers \& Graphics}, 68:1--10, 2017.

\bibitem{cao2020comprehensive}
Wenming Cao, Zhiyue Yan, Zhiquan He, and Zhihai He.
\newblock A comprehensive survey on geometric deep learning.
\newblock {\em IEEE Access}, 8:35929--35949, 2020.

\bibitem{cosmo2016shrec}
Luca Cosmo, Emanuele Rodola, Michael~M Bronstein, Andrea Torsello, Daniel
  Cremers, and Y Sahillioglu.
\newblock Shrec’16: Partial matching of deformable shapes.
\newblock {\em Proc. 3DOR}, 2(9):12, 2016.

\bibitem{Deng_2018_CVPR}
Haowen Deng, Tolga Birdal, and Slobodan Ilic.
\newblock {PPFNet}: Global context aware local features for robust 3d point
  matching.
\newblock In {\em CVPR}, 2018.

\bibitem{Dinh2005}
Huong~Quynh Dinh, Anthony Yezzi, and Greg Turk.
\newblock Texture transfer during shape transformation.
\newblock {\em {ACM} Transactions on Graphics}, 24(2):289--310, Apr. 2005.

\bibitem{donati-duo}
Nicolas Donati, Etienne Corman, and Maks Ovsjanikov.
\newblock {Deep Orientation-Aware Functional Maps: Tackling Symmetry Issues in
  Shape Matching}.
\newblock {\em {CVPR ''12 : IEEE Conference on Computer Vision and Pattern
  Recognition}}, 2022.

\bibitem{donati2020deep}
Nicolas Donati, Abhishek Sharma, and Maks Ovsjanikov.
\newblock Deep geometric functional maps: Robust feature learning for shape
  correspondence.
\newblock In {\em Proceedings of the IEEE/CVF Conference on Computer Vision and
  Pattern Recognition}, pages 8592--8601, 2020.

\bibitem{Eisenberger2020SmoothSM}
Marvin Eisenberger, Zorah L{\"a}hner, and Daniel Cremers.
\newblock Smooth shells: Multi-scale shape registration with functional maps.
\newblock {\em 2020 IEEE/CVF Conference on Computer Vision and Pattern
  Recognition (CVPR)}, pages 12262--12271, 2020.

\bibitem{Eisenberger2021NeuroMorphUS}
Marvin Eisenberger, David Novotn{\'y}, Gael Kerchenbaum, Patrick Labatut,
  Natalia Neverova, Daniel Cremers, and Andrea Vedaldi.
\newblock Neuromorph: Unsupervised shape interpolation and correspondence in
  one go.
\newblock {\em 2021 IEEE/CVF Conference on Computer Vision and Pattern
  Recognition (CVPR)}, pages 7469--7479, 2021.

\bibitem{eisenberger2020deep}
Marvin Eisenberger, Aysim Toker, Laura Leal-Taix\'{e}, and Daniel Cremers.
\newblock Deep shells: Unsupervised shape correspondence with optimal
  transport.
\newblock In H. Larochelle, M. Ranzato, R. Hadsell, M.~F. Balcan, and H. Lin,
  editors, {\em Advances in Neural Information Processing Systems}, volume~33,
  pages 10491--10502. Curran Associates, Inc., 2020.

\bibitem{eynard2016coupled}
Davide Eynard, Emanuele Rodola, Klaus Glashoff, and Michael~M Bronstein.
\newblock Coupled functional maps.
\newblock In {\em 3D Vision (3DV)}, pages 399--407. IEEE, 2016.

\bibitem{ezuz2017deblurring}
Danielle Ezuz and Mirela Ben-Chen.
\newblock Deblurring and denoising of maps between shapes.
\newblock In {\em Computer Graphics Forum}. Wiley Online Library, 2017.

\bibitem{fey2018splinecnn}
Matthias Fey, Jan~Eric Lenssen, Frank Weichert, and Heinrich M{\"u}ller.
\newblock Splinecnn: Fast geometric deep learning with continuous b-spline
  kernels.
\newblock In {\em Proceedings of the IEEE Conference on Computer Vision and
  Pattern Recognition}, pages 869--877, 2018.

\bibitem{ginzburg2019cyclic}
Dvir Ginzburg and Dan Raviv.
\newblock Cyclic functional mapping: Self-supervised correspondence between
  non-isometric deformable shapes, 2019.

\bibitem{Ginzburg2020}
Dvir Ginzburg and Dan Raviv.
\newblock Cyclic functional mapping: Self-supervised correspondence between
  non-isometric deformable shapes.
\newblock In {\em Computer Vision {\textendash} {ECCV} 2020}, pages 36--52.
  Springer International Publishing, 2020.

\bibitem{Gojcic_2019_CVPR}
Zan Gojcic, Caifa Zhou, Jan~D. Wegner, and Andreas Wieser.
\newblock The perfect match: {3D} point cloud matching with smoothed densities.
\newblock In {\em CVPR}, 2019.

\bibitem{gong2019spiralnet++}
Shunwang Gong, Lei Chen, Michael Bronstein, and Stefanos Zafeiriou.
\newblock Spiralnet++: A fast and highly efficient mesh convolution operator.
\newblock In {\em Proceedings of the IEEE/CVF International Conference on
  Computer Vision Workshops}, pages 0--0, 2019.

\bibitem{groueix20183d}
Thibault Groueix, Matthew Fisher, Vladimir~G Kim, Bryan~C Russell, and Mathieu
  Aubry.
\newblock 3d-coded: 3d correspondences by deep deformation.
\newblock In {\em Proceedings of the European Conference on Computer Vision
  (ECCV)}, pages 230--246, 2018.

\bibitem{guo2016comprehensive}
Yulan Guo, Mohammed Bennamoun, Ferdous Sohel, Min Lu, Jianwei Wan, and
  Ngai~Ming Kwok.
\newblock A comprehensive performance evaluation of 3d local feature
  descriptors.
\newblock {\em International Journal of Computer Vision}, 116(1):66--89, 2016.

\bibitem{guo2020deep}
Yulan Guo, Hanyun Wang, Qingyong Hu, Hao Liu, Li Liu, and Mohammed Bennamoun.
\newblock Deep learning for 3d point clouds: A survey.
\newblock {\em IEEE transactions on pattern analysis and machine intelligence},
  2020.

\bibitem{halimi2019unsupervised}
Oshri Halimi, Or Litany, Emanuele Rodola, Alex~M Bronstein, and Ron Kimmel.
\newblock Unsupervised learning of dense shape correspondence.
\newblock In {\em Proceedings of the IEEE/CVF Conference on Computer Vision and
  Pattern Recognition}, pages 4370--4379, 2019.

\bibitem{huang2014functional}
Qixing Huang, Fan Wang, and Leonidas Guibas.
\newblock Functional map networks for analyzing and exploring large shape
  collections.
\newblock {\em ACM Transactions on Graphics (TOG)}, 33(4):36, 2014.

\bibitem{Kim2011}
Vladimir~G. Kim, Yaron Lipman, and Thomas Funkhouser.
\newblock Blended intrinsic maps.
\newblock {\em {ACM} Transactions on Graphics}, 30(4):1--12, July 2011.

\bibitem{kingma2017adam}
Diederik~P. Kingma and Jimmy Ba.
\newblock Adam: A method for stochastic optimization, 2017.

\bibitem{kostrikov2018surface}
Ilya Kostrikov, Zhongshi Jiang, Daniele Panozzo, Denis Zorin, and Joan Bruna.
\newblock Surface networks.
\newblock In {\em Proceedings of the IEEE Conference on Computer Vision and
  Pattern Recognition}, pages 2540--2548, 2018.

\bibitem{kovnatsky2013coupled}
Artiom Kovnatsky, Michael~M Bronstein, Alexander~M Bronstein, Klaus Glashoff,
  and Ron Kimmel.
\newblock Coupled quasi-harmonic bases.
\newblock In {\em Computer Graphics Forum}, volume~32, pages 439--448. Wiley
  Online Library, 2013.

\bibitem{li2022srfeat}
Lei Li, Souhaib Attaiki, and Maks Ovsjanikov.
\newblock {SRFeat}: Learning locally accurate and globally consistent non-rigid
  shape correspondence.
\newblock In {\em 2022 International Conference on 3D Vision (3DV)}. {IEEE},
  Sept. 2022.

\bibitem{litany2017deep}
Or Litany, Tal Remez, Emanuele Rodola, Alex Bronstein, and Michael Bronstein.
\newblock Deep functional maps: Structured prediction for dense shape
  correspondence.
\newblock In {\em Proceedings of the IEEE international conference on computer
  vision}, pages 5659--5667, 2017.

\bibitem{Litany2017}
O. Litany, E. Rodol{\`{a}}, A.~M. Bronstein, and M.~M. Bronstein.
\newblock Fully spectral partial shape matching.
\newblock {\em Computer Graphics Forum}, 36(2):247--258, May 2017.

\bibitem{Litany2016}
O. Litany, E. Rodol{\`{a}}, A.~M. Bronstein, M.~M. Bronstein, and D. Cremers.
\newblock Non-rigid puzzles.
\newblock {\em Computer Graphics Forum}, 35(5):135--143, Aug. 2016.

\bibitem{litman2013learning}
Roee Litman and Alexander~M Bronstein.
\newblock Learning spectral descriptors for deformable shape correspondence.
\newblock {\em IEEE transactions on pattern analysis and machine intelligence},
  36(1):171--180, 2013.

\bibitem{rethink_ma_22}
Xu Ma, Can Qin, Haoxuan You, Haoxi Ran, and Yun Fu.
\newblock Rethinking network design and local geometry in point cloud: A simple
  residual mlp framework, 2022.

\bibitem{marin22_why}
Riccardo Marin, Souhaib Attaiki, Simone Melzi, Emanuele Rodolà, and Maks
  Ovsjanikov.
\newblock Why you should learn functional basis, 2021.

\bibitem{Marin2020CorrespondenceLV}
Riccardo Marin, Marie-Julie Rakotosaona, Simone Melzi, and Maks Ovsjanikov.
\newblock Correspondence learning via linearly-invariant embedding.
\newblock {\em ArXiv}, abs/2010.13136, 2020.

\bibitem{maron2017convolutional}
Haggai Maron, Meirav Galun, Noam Aigerman, Miri Trope, Nadav Dym, Ersin Yumer,
  Vladimir~G Kim, and Yaron Lipman.
\newblock Convolutional neural networks on surfaces via seamless toric covers.
\newblock {\em ACM Trans. Graph.}, 36(4):71--1, 2017.

\bibitem{masci2015geodesic}
Jonathan Masci, Davide Boscaini, Michael Bronstein, and Pierre Vandergheynst.
\newblock Geodesic convolutional neural networks on riemannian manifolds.
\newblock In {\em Proceedings of the IEEE international conference on computer
  vision workshops}, pages 37--45, 2015.

\bibitem{shrec19}
S. Melzi, R. Marin, E. Rodolà, U. Castellani, J. Ren, A. Poulenard, P. Wonka,
  and M. Ovsjanikov.
\newblock Matching humans with different connectivity.
\newblock {\em Eurographics Workshop on 3D Object Retrieval}, 2019.

\bibitem{Melzi_2019}
Simone Melzi, Jing Ren, Emanuele Rodolà, Abhishek Sharma, Peter Wonka, and
  Maks Ovsjanikov.
\newblock {Z}oom{O}ut: Spectral upsampling for efficient shape correspondence.
\newblock {\em ACM Transactions on Graphics}, 38(6):1–14, Nov 2019.

\bibitem{Meyer2003}
Mark Meyer, Mathieu Desbrun, Peter Schr\"{o}der, and Alan~H. Barr.
\newblock Discrete differential-geometry operators for triangulated
  2-manifolds.
\newblock In {\em Mathematics and Visualization}, pages 35--57. Springer Berlin
  Heidelberg, 2003.

\bibitem{monti2017geometric}
Federico Monti, Davide Boscaini, Jonathan Masci, Emanuele Rodola, Jan Svoboda,
  and Michael~M Bronstein.
\newblock Geometric deep learning on graphs and manifolds using mixture model
  cnns.
\newblock In {\em Proceedings of the IEEE conference on computer vision and
  pattern recognition}, pages 5115--5124, 2017.

\bibitem{Nogneng2017}
Dorian Nogneng and Maks Ovsjanikov.
\newblock Informative descriptor preservation via commutativity for shape
  matching.
\newblock {\em Computer Graphics Forum}, 36(2):259--267, May 2017.

\bibitem{Ovsjanikov2012}
Maks Ovsjanikov, Mirela Ben-Chen, Justin Solomon, Adrian Butscher, and Leonidas
  Guibas.
\newblock Functional maps.
\newblock {\em {ACM} Transactions on Graphics}, 31(4):1--11, Aug. 2012.

\bibitem{ovsjanikov2012functional}
Maks Ovsjanikov, Mirela Ben-Chen, Justin Solomon, Adrian Butscher, and Leonidas
  Guibas.
\newblock {F}unctional {M}aps: {A} {F}lexible {R}epresentation of {M}aps
  {B}etween {S}hapes.
\newblock {\em ACM Transactions on Graphics (TOG)}, 31(4):30, 2012.

\bibitem{Ovsjanikov2017}
Maks Ovsjanikov, Etienne Corman, Michael Bronstein, Emanuele Rodol{\`{a}},
  Mirela Ben-Chen, Leonidas Guibas, Frederic Chazal, and Alex Bronstein.
\newblock Computing and processing correspondences with functional maps.
\newblock In {\em {ACM} {SIGGRAPH} 2017 Courses}. {ACM}, July 2017.

\bibitem{Pai_2021_CVPR}
Gautam Pai, Jing Ren, Simone Melzi, Peter Wonka, and Maks Ovsjanikov.
\newblock Fast sinkhorn filters: Using matrix scaling for non-rigid shape
  correspondence with functional maps.
\newblock In {\em Proceedings of the IEEE/CVF Conference on Computer Vision and
  Pattern Recognition (CVPR)}, pages 384--393, June 2021.

\bibitem{Pishchulin2017}
Leonid Pishchulin, Stefanie Wuhrer, Thomas Helten, Christian Theobalt, and
  Bernt Schiele.
\newblock Building statistical shape spaces for 3d human modeling.
\newblock {\em Pattern Recognition}, 67:276--286, July 2017.

\bibitem{poulenard2018multi}
Adrien Poulenard and Maks Ovsjanikov.
\newblock Multi-directional geodesic neural networks via equivariant
  convolution.
\newblock {\em ACM Transactions on Graphics (TOG)}, 37(6):1--14, 2018.

\bibitem{poulenard2019effective}
Adrien Poulenard, Marie-Julie Rakotosaona, Yann Ponty, and Maks Ovsjanikov.
\newblock Effective rotation-invariant point cnn with spherical harmonics
  kernels.
\newblock In {\em 2019 International Conference on 3D Vision (3DV)}, pages
  47--56. IEEE, 2019.

\bibitem{poulenard_persistence}
Adrien Poulenard, Primoz Skraba, and Maks Ovsjanikov.
\newblock {Topological Function Optimization for Continuous Shape Matching}.
\newblock {\em {Computer Graphics Forum}}, 37(5):13--25, 2018.

\bibitem{qi2017pointnet}
Charles~R Qi, Hao Su, Kaichun Mo, and Leonidas~J Guibas.
\newblock Pointnet: Deep learning on point sets for 3d classification and
  segmentation.
\newblock {\em Proc. Computer Vision and Pattern Recognition (CVPR), IEEE},
  1(2):4, 2017.

\bibitem{qi2017pointnet++}
Charles~Ruizhongtai Qi, Li Yi, Hao Su, and Leonidas~J Guibas.
\newblock Pointnet++: Deep hierarchical feature learning on point sets in a
  metric space.
\newblock In {\em Advances in Neural Information Processing Systems}, pages
  5105--5114, 2017.

\bibitem{jing_maptree}
Jing Ren, Simone Melzi, Maks Ovsjanikov, and Peter Wonka.
\newblock Maptree.
\newblock {\em ACM Transactions on Graphics}, 39(6):1–17, Dec 2020.

\bibitem{discrete_Ren2021}
Jing Ren, Simone Melzi, Peter Wonka, and Maks Ovsjanikov.
\newblock Discrete optimization for shape matching.
\newblock {\em Computer Graphics Forum}, 40(5):81--96, Aug. 2021.

\bibitem{ren2021discrete}
Jing Ren, Simone Melzi, Peter Wonka, and Maks Ovsjanikov.
\newblock Discrete optimization for shape matching.
\newblock In {\em Computer Graphics Forum}. Wiley Online Library, 2021.

\bibitem{Ren2019}
Jing Ren, Adrien Poulenard, Peter Wonka, and Maks Ovsjanikov.
\newblock Continuous and orientation-preserving correspondences via functional
  maps.
\newblock {\em ACM Transactions on Graphics (ToG)}, 37(6):1--16, 2018.

\bibitem{ren2018continuous}
Jing Ren, Adrien Poulenard, Peter Wonka, and Maks Ovsjanikov.
\newblock Continuous and orientation-preserving correspondences via functional
  maps.
\newblock {\em ACM Transactions on Graphics (TOG)}, 37(6), 2018.

\bibitem{Rodol2016}
E. Rodol{\`{a}}, L. Cosmo, M.~M. Bronstein, A. Torsello, and D. Cremers.
\newblock Partial functional correspondence.
\newblock {\em Computer Graphics Forum}, 36(1):222--236, Feb. 2016.

\bibitem{rodola2017partial}
Emanuele Rodol{\`a}, Luca Cosmo, Michael~M Bronstein, Andrea Torsello, and
  Daniel Cremers.
\newblock Partial functional correspondence.
\newblock In {\em Computer Graphics Forum}, volume~36, pages 222--236. Wiley
  Online Library, 2017.

\bibitem{roufosse2019unsupervised}
Jean-Michel Roufosse, Abhishek Sharma, and Maks Ovsjanikov.
\newblock Unsupervised deep learning for structured shape matching.
\newblock In {\em Proceedings of the IEEE/CVF International Conference on
  Computer Vision}, pages 1617--1627, 2019.

\bibitem{rustamov2013map}
Raif~M Rustamov, Maks Ovsjanikov, Omri Azencot, Mirela Ben-Chen,
  Fr{\'e}d{\'e}ric Chazal, and Leonidas Guibas.
\newblock Map-based exploration of intrinsic shape differences and variability.
\newblock {\em ACM Transactions on Graphics (TOG)}, 32(4):1--12, 2013.

\bibitem{Salti2014}
Samuele Salti, Federico Tombari, and Luigi~Di Stefano.
\newblock {SHOT}: Unique signatures of histograms for surface and texture
  description.
\newblock {\em Computer Vision and Image Understanding}, 125:251--264, Aug.
  2014.

\bibitem{sharma2020weakly}
Abhishek Sharma and Maks Ovsjanikov.
\newblock Weakly supervised deep functional maps for shape matching.
\newblock In H. Larochelle, M. Ranzato, R. Hadsell, M.~F. Balcan, and H. Lin,
  editors, {\em Advances in Neural Information Processing Systems}, volume~33,
  pages 19264--19275. Curran Associates, Inc., 2020.

\bibitem{sharp2021diffusion}
Nicholas Sharp, Souhaib Attaiki, Keenan Crane, and Maks Ovsjanikov.
\newblock Diffusionnet: Discretization agnostic learning on surfaces.
\newblock {\em ACM Trans. Graph.}, 01(1), 2022.

\bibitem{sharp2020laplacian}
Nicholas Sharp and Keenan Crane.
\newblock A laplacian for nonmanifold triangle meshes.
\newblock In {\em Computer Graphics Forum}, volume~39, pages 69--80. Wiley
  Online Library, 2020.

\bibitem{Shoham2019}
Meged Shoham, Amir Vaxman, and Mirela Ben-Chen.
\newblock Hierarchical functional maps between subdivision surfaces.
\newblock {\em Computer Graphics Forum}, 38(5):55--73, Aug. 2019.

\bibitem{sun2009concise}
Jian Sun, Maks Ovsjanikov, and Leonidas Guibas.
\newblock A concise and provably informative multi-scale signature based on
  heat diffusion.
\newblock In {\em Computer graphics forum}, pages 1383--1392. Wiley Online
  Library, 2009.

\bibitem{thomas2019KPConv}
Hugues Thomas, Charles~R. Qi, Jean-Emmanuel Deschaud, Beatriz Marcotegui,
  Fran{\c{c}}ois Goulette, and Leonidas~J. Guibas.
\newblock Kpconv: Flexible and deformable convolution for point clouds.
\newblock {\em Proceedings of the IEEE International Conference on Computer
  Vision}, 2019.

\bibitem{trappolini2021shape}
Giovanni Trappolini, Luca Cosmo, Luca Moschella, Riccardo Marin, Simone Melzi,
  and Emanuele Rodol{\`a}.
\newblock Shape registration in the time of transformers.
\newblock {\em Advances in Neural Information Processing Systems},
  34:5731--5744, 2021.

\bibitem{dgcnn}
Yue Wang, Yongbin Sun, Ziwei Liu, Sanjay~E. Sarma, Michael~M. Bronstein, and
  Justin~M. Solomon.
\newblock Dynamic graph cnn for learning on point clouds.
\newblock {\em ACM Transactions on Graphics (TOG)}, 2019.

\bibitem{wiersma2020cnns}
Ruben Wiersma, Elmar Eisemann, and Klaus Hildebrandt.
\newblock Cnns on surfaces using rotation-equivariant features.
\newblock {\em ACM Transactions on Graphics (TOG)}, 39(4):92--1, 2020.

\bibitem{Wiersma2022DeltaConv}
Ruben Wiersma, Ahmad Nasikun, Elmar Eisemann, and Klaus Hildebrandt.
\newblock {DeltaConv}.
\newblock {\em {ACM} Transactions on Graphics}, 41(4):1--10, July 2022.

\bibitem{Wu2020}
Yan Wu, Jun Yang, and Jinlong Zhao.
\newblock Partial 3d shape functional correspondence via fully spectral
  eigenvalue alignment and upsampling refinement.
\newblock {\em Computers {\&} Graphics}, 92:99--113, Nov. 2020.

\bibitem{Xiang_2021_CVPR}
Rui Xiang, Rongjie Lai, and Hongkai Zhao.
\newblock A dual iterative refinement method for non-rigid shape matching.
\newblock In {\em Proceedings of the IEEE/CVF Conference on Computer Vision and
  Pattern Recognition (CVPR)}, pages 15930--15939, June 2021.

\bibitem{Yew_2020_CVPR}
Zi~Jian Yew and Gim~Hee Lee.
\newblock {RPM-Net}: Robust point matching using learned features.
\newblock In {\em CVPR}, 2020.

\bibitem{Zeng_2017_CVPR}
Andy Zeng, Shuran Song, Matthias Niessner, Matthew Fisher, Jianxiong Xiao, and
  Thomas Funkhouser.
\newblock {3DMatch}: Learning local geometric descriptors from rgb-d
  reconstructions.
\newblock In {\em CVPR}, 2017.

\bibitem{Zhou2016}
Qian-Yi Zhou, Jaesik Park, and Vladlen Koltun.
\newblock Fast global registration.
\newblock In {\em Computer Vision {\textendash} {ECCV} 2016}, pages 766--782.
  Springer International Publishing, 2016.

\bibitem{Zuffi:CVPR:2017}
Silvia Zuffi, Angjoo Kanazawa, David Jacobs, and Michael~J. Black.
\newblock {3D} menagerie: Modeling the {3D} shape and pose of animals.
\newblock In {\em IEEE Conf. on Computer Vision and Pattern Recognition
  (CVPR)}, July 2017.

\end{thebibliography}
}

\end{document}